\documentclass[runningheads]{llncs}

\usepackage{graphicx}
\usepackage{amsfonts}
\usepackage{amsmath}
\usepackage{booktabs}
\usepackage{xcolor}
\usepackage[misc]{ifsym}
\usepackage{hyperref}

\usepackage{placeins}

\newcommand{\corr}{(\Letter)}
\usepackage{mwe}

\def\our{TACTIC}
\usepackage{subcaption}

\begin{document}



\title{TACTIC for Navigating the Unknown: Tabular Anomaly deteCTion via In-Context inference}

\titlerunning{TACTIC for Navigating the Unknown}

\author{Patryk Marsza{\l}ek\inst{1, 2} \corr \and Tomasz Ku{\'s}mierczyk\inst{1} \and Marek {\'S}mieja\inst{1}}

\authorrunning{P. Marsza{\l}ek}

\institute{Faculty of Mathematics and Computer Science, \\Jagiellonian University, Kraków, Poland \and Doctoral School of Exact and Natural Sciences, \\Jagiellonian University, Kraków, Poland \\ \vspace{0.2cm} \email{patryk.marszalek@doctoral.uj.edu.pl} \\ \email{tomasz.kusmierczyk@gmail.com} \\ \email{marek.smieja@uj.edu.pl}}

\maketitle              

\begin{abstract}
Anomaly detection for tabular data has been a long-standing unsupervised learning problem that remains a major challenge for current deep learning models.
Recently, in-context learning has emerged as a new paradigm that has shifted efforts from task-specific optimization to large-scale pretraining aimed at creating foundation models that generalize across diverse datasets. Although in-context models, such as TabPFN, perform well in supervised problems, their learned classification-based priors may not readily extend to anomaly detection.

In this paper, we study in-context models for anomaly detection and show that the unsupervised extensions to TabPFN exhibit unstable behavior, particularly in noisy or contaminated contexts, in addition to the high computational cost. We address these challenges and introduce \our{}, an in-context anomaly detection approach based on pretraining with anomaly-centric synthetic priors, which provides fast and data-dependent reasoning about anomalies while avoiding dataset-specific tuning. In contrast to typical score-based approaches, which produce uncalibrated anomaly scores that require post-processing (e.g. threshold selection or ranking heuristics), the proposed model is trained as a discriminative predictor, enabling unambiguous anomaly decisions in a single forward pass.

Through experiments on real-world datasets, we examine the performance of \our{} in clean and noisy contexts with varying anomaly rates and different anomaly types, as well as the impact of prior choices on detection quality. Our experiments clearly show that specialized \linebreak anomaly-centric in-context models such as \our{} are highly competitive compared to other task-specific methods.
\keywords{Anomaly detection  \and In-context Learning \and Tabular data \and Prior Pretraining.}
\end{abstract}

\section{Introduction}

\begin{figure}[t]
    \begin{subfigure}[b]{0.48\textwidth}
        \includegraphics[width=\textwidth]{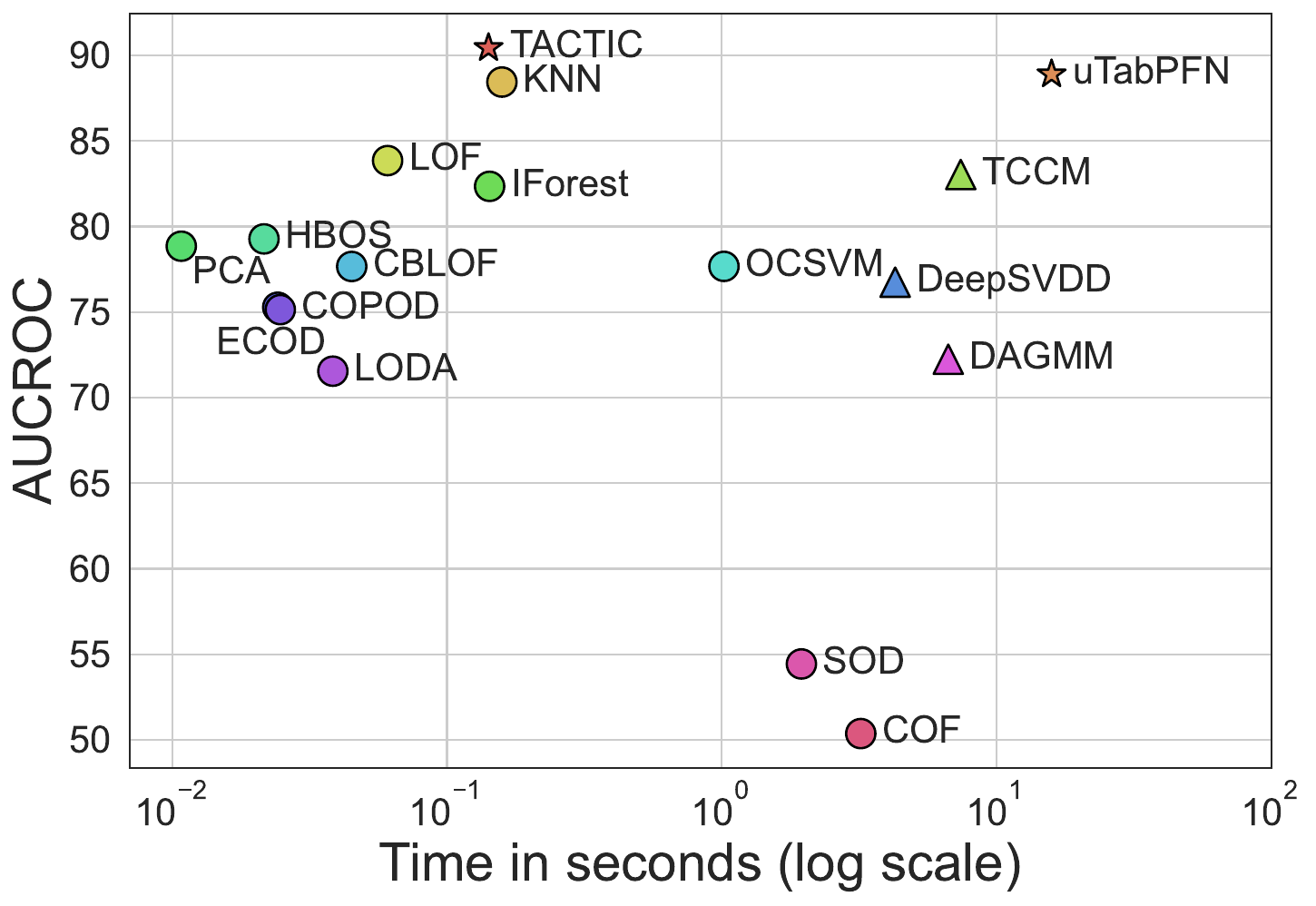}
        \caption{Clean context setting (i.e., no anomalies in reference set).}
    \end{subfigure}
    \hfill
    \begin{subfigure}[b]{0.48\textwidth}
        \includegraphics[width=\textwidth]{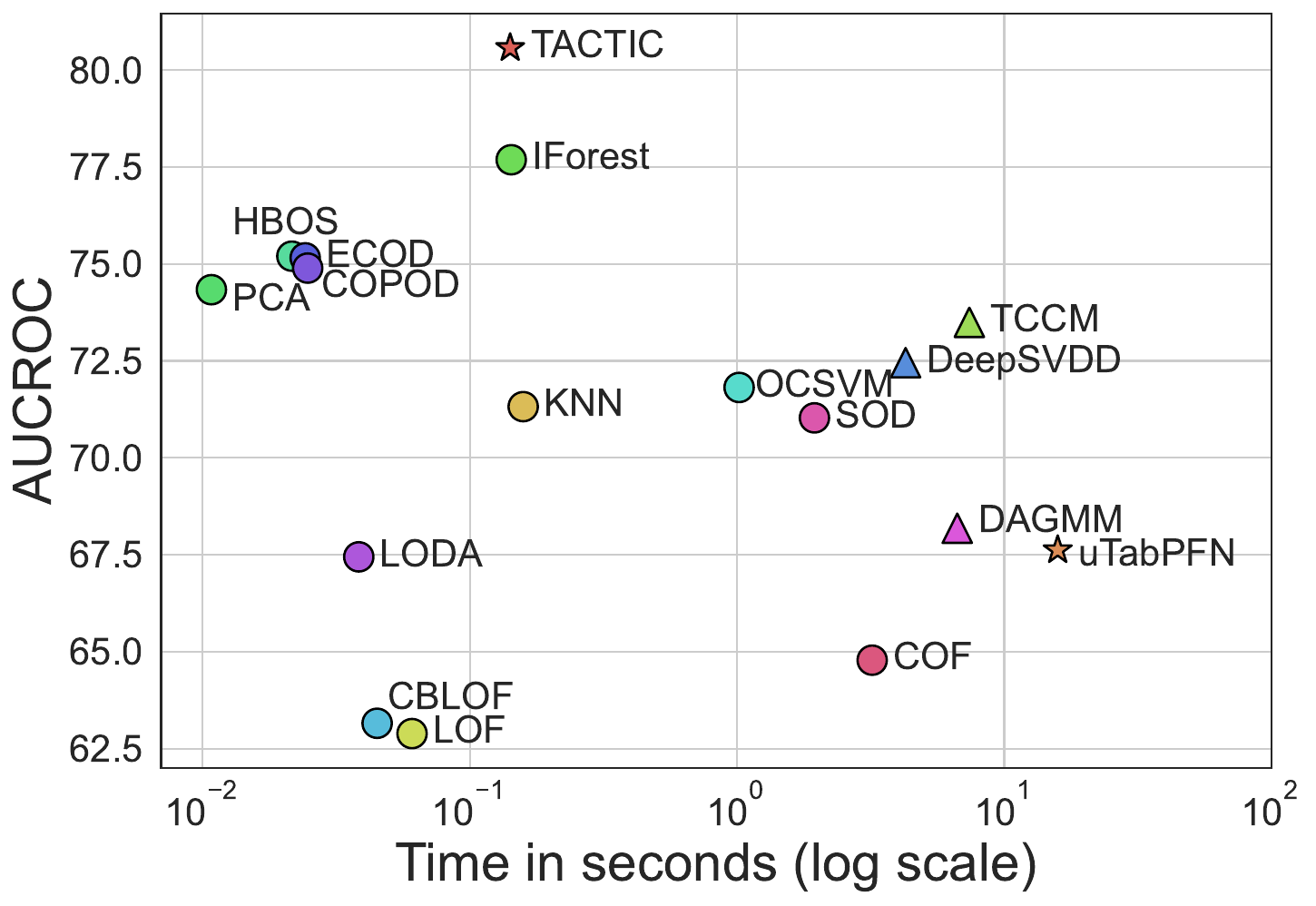}
        \caption{Noisy context setting (i.e., reference set corrupted with anomalies).}
    \end{subfigure}
    \caption{
Average AUCROC performance (vertical axis) versus computation time (horizontal axis), calculated over multiple datasets, for classical (circles), deep-learning (triangles), and in-context methods (stars). Note that the horizontal axis corresponds to inference time for out-of-the-box models and to the combined fitting and inference time for models that require per-dataset training.
In terms of evaluation time, \our{} falls within the group of fast shallow methods while achieving top performance for both clean and noisy context settings.    
    }
    \label{fig:Teaser}
\end{figure}


Anomaly detection (e.g., novelty or outlier)  plays a critical role in high-stakes domains such as cybersecurity, industrial monitoring, and healthcare, where rare and unexpected events can have severe consequences~\cite{han2022adbench}.
It refers to the identification of abnormal or novel patterns embedded in nominal (normal) data. In a typical unsupervised setting, we have no access to labeled anomalies 
and anomalous observations may be mixed into the nominal training data, which makes the problem particularly challenging. 



Typical solutions addressing this problem focus on modeling nominal data and contrasting samples against the learned model~\cite{grunau2022adapting}. 
Popular examples include density estimators~\cite{an2015variational}, autoencoder-based models~\cite{zhou2017anomaly}, and one-class classifiers~\cite{maziarka2021oneflow}. Although simpler shallow methods may lack sufficient expressivity to capture complex patterns, more modern deep learning solutions are often sensitive to hyperparameter selection, slower, and require dataset-specific tuning. Additionally, most of these techniques only provide anomaly scores and therefore require post-processing to obtain final decisions.


In contrast, recently emerging in-context learning approaches have substantially changed the way tabular data is handled~\cite{tabicl,hollmann2025accurate,den2024fine}. Models such as \linebreak TabPFN~\cite{tabpfn} demonstrate that large-scale pretraining combined with in-context inference can deliver strong performance without task-specific retraining.
In particular, an unsupervised extension of TabPFN\footnote{\url{https://docs.priorlabs.ai/capabilities/anomaly-detection}} handles anomaly detection out of the box via autoregressive density estimation, which naturally yields an anomaly score. However, our experiments show that this approach incurs slow inference, especially in high-dimensional settings (Figure~\ref{fig:Teaser}), may exhibit unstable behavior due to priors learned for a different objective (e.g., classification; see Figure~\ref{fig:0_20_anomaly_level}), and does not provide a principled threshold for making decisions separating nominal samples from anomalous ones.

In this work, we introduce \our{}, an alternative in-context approach specifically tailored to the anomaly detection problem, substantially different from the unsupervised extension to TabPFN. \our{} is a discriminative model that returns already calibrated probabilities of samples being \emph{anomalous}, eliminating the need for post-processing or thresholding of anomaly scores. Given a context containing unlabeled training data, it directly solves the anomaly detection problem for all the query samples at once in a single forward pass, without the need for any additional adaptation. 
Figure~\ref{fig:Teaser} compares its superior performance and fast computation time 
against shallow, deep and in-context baselines.

Although \our{} operates in a fully unsupervised setting at inference time (i.e., no labels are provided in the context), it is pretrained in a supervised fashion using an anomaly-centric prior. 
The prior was designed to represent a wide range of possible datasets. In particular, we employ a mixture of synthetic priors, combining TabPFN-like generative processes with newly designed Gaussian mixture models augmented with specially designed anomalies. Additionally, by contaminating the context during pretraining, we make \our{} robust to realistic problem settings.


We conducted a series of experiments to highlight the key properties of \our{}. We begin by comparing our method with the TabPFN2.5 anomaly detection extension and standard benchmark methods on real-world anomaly datasets, considering both clean-context settings and settings in which anomalies are present in the context. We then systematically investigated robustness to noisy contexts by varying the proportion of anomalies in the context and test sets, both jointly and independently. Finally, we provide a focused analysis of our approach by comparing different prior choices and quantifying their impact on detection results, shedding light on how pretraining assumptions shape anomaly detection behavior.

Taken together, our study shows that while general in-context foundation models can perform well in idealized conditions, a task-focused, pretrained approach offers a more stable, efficient, and interpretable alternative for anomaly detection in realistic and noisy settings. The codebase used in this paper can be found at \href{https://github.com/gmum/TACTIC}{https://github.com/gmum/TACTIC}.

\section{\our{} for In-Context Anomaly Detection}

\begin{figure}[t!]
    \centering
    \includegraphics[width=0.7\linewidth]{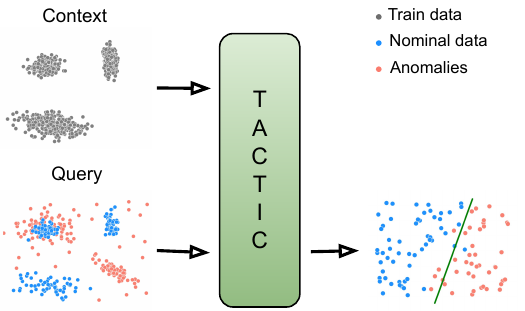}
    \caption{
Schematic overview of our approach. 
By making use of contextual information about data distribution, \our{} separates query points into nominal and anomalous.
    }
    \label{fig:method}
\end{figure}




We formulate anomaly detection as an \emph{in-context problem}. Given a dataset $D = \{x_i\}_{i=1}^n \subset \mathbb{R}^d$, we consider a \emph{context set} $D_{\text{ctx}} \subset D$ and a set of \emph{query points} $D_{\text{q}} = D \setminus D_{\text{ctx}}$. A context represents training data that are unlabeled. Based on the information contained in the context, the model has to predict whether an individual query point represents a nominal or anomalous observation. 

The proposed \our{} model directly focuses on modeling the probability 
$$
\hat p_\theta(y=\text{anomaly} \mid x, D_{\text{ctx}})
$$ 
that a query point $x$ is an anomaly or a nominal instance. The prediction is conditioned on the features of $x$ and the entire context $D_{\text{ctx}}$. This formulation defines a purely discriminative model that outputs a binary decision rather than an anomaly score, which may not be straightforward to binarize. \our{} performs predictions for all query points simultaneously in a single forward pass, which is extremely effective in practice. Once the model is pretrained, 
we freeze its weights, and at inference
no dataset-specific optimization is ever required.


The in-context formulation requires \our{} to infer what constitutes \emph{normality} solely from the structure of the observed context. This requirement is addressed by learning suitable model parameters $\theta^\star$ during pretraining. The model is trained on synthetic datasets $D \sim p(D)$ using a cross-entropy loss $\ell$ with ground-truth labels $y \in \{\text{nominal}, \text{anomaly}\}$:
\[
\theta^\star = \arg\min_\theta 
\; \mathbb{E}_{D \sim p(D)} 
\Big[ \sum_{x \in D_{\text{q}}}
\ell\big(y(x), \hat p_\theta(y=\text{anomaly} \mid x, D_{\text{ctx}})\big)
\Big],
\]
where $p(D)$ denotes the data-generating prior. This objective induces a dataset-level prior over anomaly structure: rather than learning a fixed scoring rule, the model learns how to infer anomaly patterns from contextual information.

The introduced \our{} model instantiates the above approach using a tabular transformer architecture~\cite{den2024fine}  to be capable of processing sets and jointly attending over all points in $D_{\text{ctx}} \cup D_{\text{q}}$. Each data point (table row) is linearly embedded in a 512-dimensional token representation. The model then applies 12 self-attention layers, each with 4 attention heads and a GeLU activation function, enabling it to capture global relationships such as cluster structure, density variations, and relative isolation. Similar to other tabular transformers, in the \our{} framework, context tokens attend to each other, while query tokens attend exclusively to the context. After the transformer encoder, each query token is passed through an MLP decoder with a single hidden layer of dimension 2048, which projects the representation down to a 2-dimensional output space.


From a Bayesian perspective, the proposed approach can be interpreted as amortized posterior predictive inference under a learned prior~\cite{mullertransformers}. The synthetic data-generating distribution $p(D)$ specifies a prior over datasets and anomaly-generating mechanisms, while conditioning on the context $D_{\text{ctx}}$ corresponds to Bayesian updating. During pretraining, the model learns to approximate the posterior predictive distribution
$p(y \mid x, D_{\text{ctx}})$ induced by this prior, but does so directly, without explicitly representing latent variables or performing iterative inference. This training objective is closely related to prior-data fitting in Prior-Data Fitted Networks (PFNs), where a transformer trained on samples from a task prior is shown to approximate Bayesian posterior prediction in a single forward pass~\cite{mullertransformers}. In this view, in-context inference replaces explicit Bayesian computation with a learned inference operator, enabling fast, dataset-adaptive predictions while preserving the inductive biases of the prior.

Besides using a different prior, the key difference from TabPFN-style methods concerns the use of labels in the context. In TabPFN, the context consists of tokens composed of feature vectors concatenated with class labels. Consequently, also at inference, it conditions on both inputs and observed class labels. 
In contrast, in \our{}, tokens are solely feature vectors and anomaly labels are used in the loss function to supervise the model. Consequently, \our{} does not use labels in the context at the inference stage, i.e., $D_{\text{ctx}}$ contains only unlabeled observations. 
Furthermore, to more closely resemble real-life scenarios, both during pretraining and at inference, the context may include unlabeled anomalies. However, since the number of these anomalies is much lower than the nominal data, the model can infer what constitutes normality directly from the structure of the context. 
From a Bayesian perspective, this procedure corresponds to amortized inference with partially observed data, where anomaly labels are treated as latent variables and are only predicted for query points. 
Overall, this setting clearly distinguishes our approach from classification-based PFN models.



\section{Prior Synthetic Data for Anomaly Detection}
\label{sec:priors}

\begin{figure}[t!]
    \centering
    \includegraphics[width=0.95\linewidth]{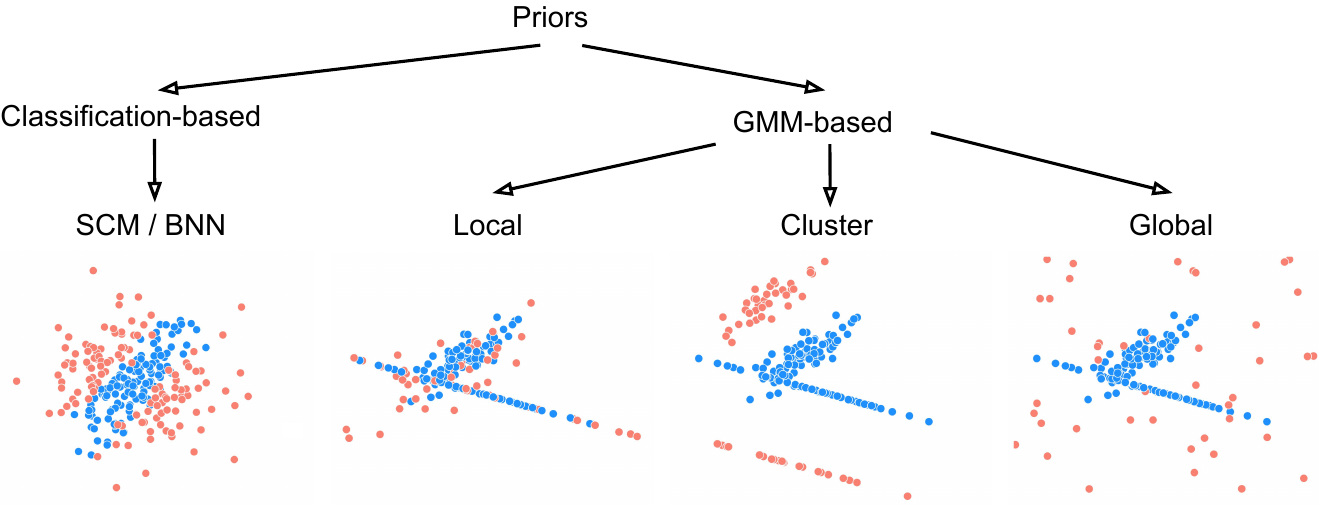}
    \caption{Anomaly-centric priors used for pretraining \our{} (nominal data is marked in blue; anomalies marked in red). Using a mixture of classification-based and GMM-based priors with various anomaly types improves the generalization ability and robustness of our method.}
    \label{fig:priors}
\end{figure}

Pretraining plays a central role in our approach, as it establishes the inductive biases of the model before it observes any real data. In contrast to conventional anomaly detection methods, which are typically fitted separately for each dataset, our model relies exclusively on knowledge acquired during an offline pretraining phase. The design of the pretraining procedure determines the implicit notions of \emph{normality} and \emph{anomalousness}, its robustness to noise in the context, and its ability to generalize to real-world datasets. Consequently, identifying an effective pretraining design constitutes a key challenge.

The 
prior $p(D)$ used during pretraining does not impose a single fixed definition of anomalous behavior. Following previous work~\cite{han2022adbench,steinbuss2021benchmarking}, we consider three well-established types of anomalies that commonly occur in real-world scenarios: local, global, and cluster anomalies. Since these categories may not capture all forms of abnormal behavior, we additionally simulate a wide spectrum of anomalies using standard classification datasets, where a subset of classes is labeled as anomalies. This results in a mixture of synthetic and classification priors (see Figure~\ref{fig:priors}), which are sampled during the pretraining phase with probabilities $0.3$ and $0.7$, respectively. Below, we describe these priors in detail:


\textbf{Prior Type I (TabPFN-based): classification-based prior.}
Following the standard TabPFN generation mechanism using Structural Causal Models (SCMs) and Bayesian Neural Networks (BNNs), we generate a dataset with 
$C \sim \mathcal{U}\{2,\dots,$ $10\}$ classes and sample a subset of \emph{nominal} classes $\mathcal{C}_{\mathrm{nom}} \subset \{1,\dots,C\}$. Samples from classes in $\mathcal{C}_{\mathrm{nom}}$ are labeled \textit{nominal}, while samples from the complementary set
$\mathcal{C}_{\mathrm{anom}} = \{1,\dots,C\} \setminus \mathcal{C}_{\mathrm{norm}}$
are labeled \textit{anomalous}. This induces anomalies at the level of entire data modes rather than local perturbations.

\textbf{Prior Type II: (GMM-based) synthetic ADBench anomalies.}
The second prior type is based on an explicit probabilistic model of nominal data, augmented with controlled mechanisms for anomaly generation. We assume that the nominal observations come from a Gaussian mixture model $\sum_{m=1}^M \pi_m \, \mathcal{N}(x \mid \mu_m, \Sigma_m)$, where the number of components $M$ is sampled from $\mathcal{U}\{1, \dots, 20\}$.
Anomalous observations are subsequently generated and added to the nominal data using one of three mechanisms inspired by AdBench~\cite{han2022adbench,steinbuss2021benchmarking}, which represent distinct types of anomalies that appear in real-life problems. Each of the following mechanisms has the same chance of being used:
\begin{itemize}
    \item \emph{Local} anomalies: component covariances are inflated according to
    $
    \Sigma_m^{\text{anom}} = \alpha \Sigma_m
    $ (with $\alpha = 5$)
    and anomalous points are sampled from $\mathcal{N}(\mu_m, \Sigma_m^{\text{anom}})$. This produces observations that remain close to high-density regions, while violating local concentration properties.

    \item \emph{Cluster} anomalies: component means are shifted as
    $
    \mu_m^{\text{anom}} = \alpha \mu_m
    $  (with $\alpha = 5$)
    while keeping the covariance fixed. This results in coherent anomalous clusters that are well separated from the nominal data manifold.

    \item \emph{Global} anomalies: after rescaling each feature of the nominal data to the interval $[-1,1]$, anomalous points are sampled from a uniform distribution
    $
    x \sim \mathcal{U}(-1.1, 1.1)^d,
    $ where $d$ is the input dimensionality.
    These samples violate global support constraints and correspond to extreme outliers.
\end{itemize}

To avoid specialization to a fixed feature space, the dimensionality of the data $d$ is also randomized. 
For each 
dataset $D \sim p(D)$, it is sampled as
$
d \sim \mathcal{U}\{2,\dots,50\},
$
and all observations are generated in $\mathbb{R}^d$. Exposure to varying dimensionalities during pretraining improves robustness and generalization.

The model is trained on datasets $D \sim p(D)$, split into context and query as $(D_{\text{ctx}}, D_{\text{q}})$. 
The training splits emulate the distinction between known nominal data (=context) and test data (=query) at inference.
The query set is constructed to contain equal numbers of nominal and anomalous samples, while the context is generated according to one of two protocols: \textbf{(C) a clean setting} with no anomalies, or \textbf{(N) a contaminated setting} in which the number of anomalies is sampled uniformly from $[0.05, 0.3]$ times the number of nominal points. Explicitly varying the proportion of anomalies in the context during pretraining encourages robustness and prevents collapse when anomalies are present in $D_{\text{ctx}}$.

In summary, variability in the prior is introduced across data-generating mechanisms, feature dimensionalities, latent structures, anomaly types, anomaly rates, and context contamination levels.
Training under such a complex prior encourages the model to develop a flexible notion of anomalies relative to data-generating processes rather than to a single dataset, consistent with the PFN paradigm of approximate Bayesian inference. 
It generalizes well across domains and dataset sizes, while remaining computationally efficient compared to iterative density estimation or likelihood-based baselines.


\section{Empirical Study}

We conducted a systematic empirical investigation into pretraining in-context models for anomaly detection. Our goal was to understand how the large-scale pretraining paradigm combined with in-context inference behaves in realistic settings, how robust it is to noisy data, and how sensitive it is to design choices in the pretraining process. 

We studied two variants of our model pretrained with clean and noisy contexts and analyze their behavior on both real and synthetic datasets. For pretraining we employed Adam optimizer with a cosine scheduler and a learning rate of 0.0001 over 120000 epochs. The batch size was set to 4 with 16 gradient accumulation steps (effective batch size of 64).
For baselines we employed  the dominant 14 algorithms from the ADBench benchmark, implemented using the PyOD~\cite{pyod} library, including: KNN~\cite{knn}, LOF~\cite{lof}, IForest~\cite{isoforest}, PCA~\cite{pca}, HBOS~\cite{hbos},  OCSVM~\cite{ocsvm}, CBLOF~\cite{cblof}, DeepSVDD~\cite{deepsvdd}, ECOD~\cite{ecod}, COPOD~\cite{copod}, LODA~\cite{loda}, SOD~\cite{sod}, COF~\cite{cof} and DAGMM~\cite{zong2018deep}. We extended the comparison by including also two more recent methods, namely the unsupervised extension of TabPFN2.5 and TCCM~\cite{tccm}. For all of the methods mentioned above, we used the hyperparameters recommended by their authors.


We compare the performance of the methods studied using AUCROC and their corresponding ranks across 38 real-world datasets from ADBench benchmark~\cite{han2022adbench}, each containing up to 50 features (see supplementary materials for details). We adhered to the original ADBench experimental pipeline. Specifically, datasets exceeding 10000 instances were randomly subsampled, whereas datasets with fewer than 1000 instances were oversampled. In addition, all features were subsequently normalized to the range $[-1,1]$. Whenever a method provides a clear decision or threshold for determining whether a sample is anomalous, we also report the F1 score and the respective ranks. To ensure stability, each experiment was averaged over 5 distinct random seeds.

{The main text contains a summary of each experiment, while detailed numerical results are provided in the supplementary materials.}

\subsection{Can We Pretrain for Real Anomaly Detection?}
\label{sec:clean_context}

\begin{figure}[htpb!]
    \centering
    
    \begin{subfigure}[b]{0.44\textwidth}
        \includegraphics[height=2.5cm]{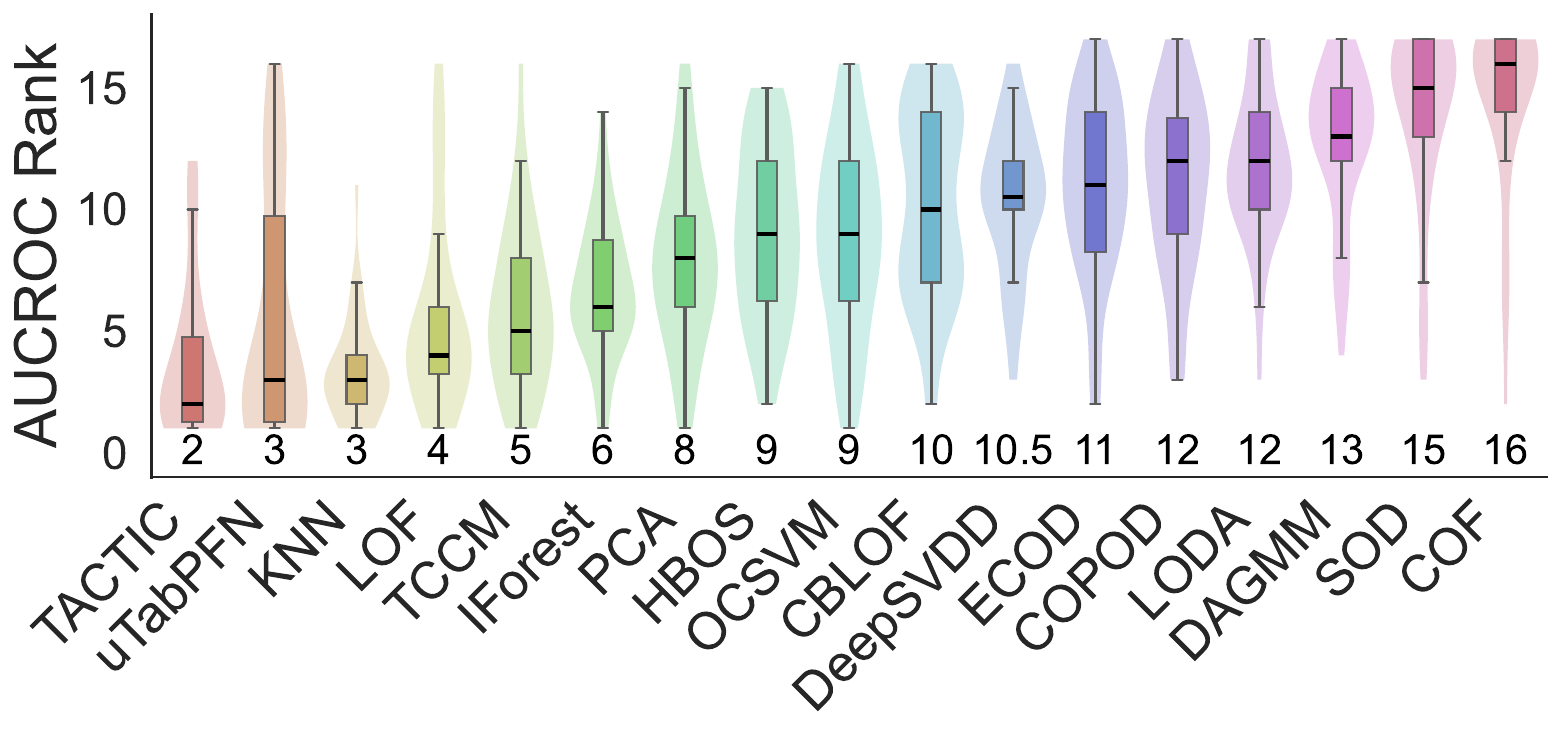}
        \caption{Distribution of AUCROC ranks.}
    \end{subfigure}
    \hfill
    \begin{subfigure}[b]{0.55\textwidth}
        \includegraphics[height=2.5cm]{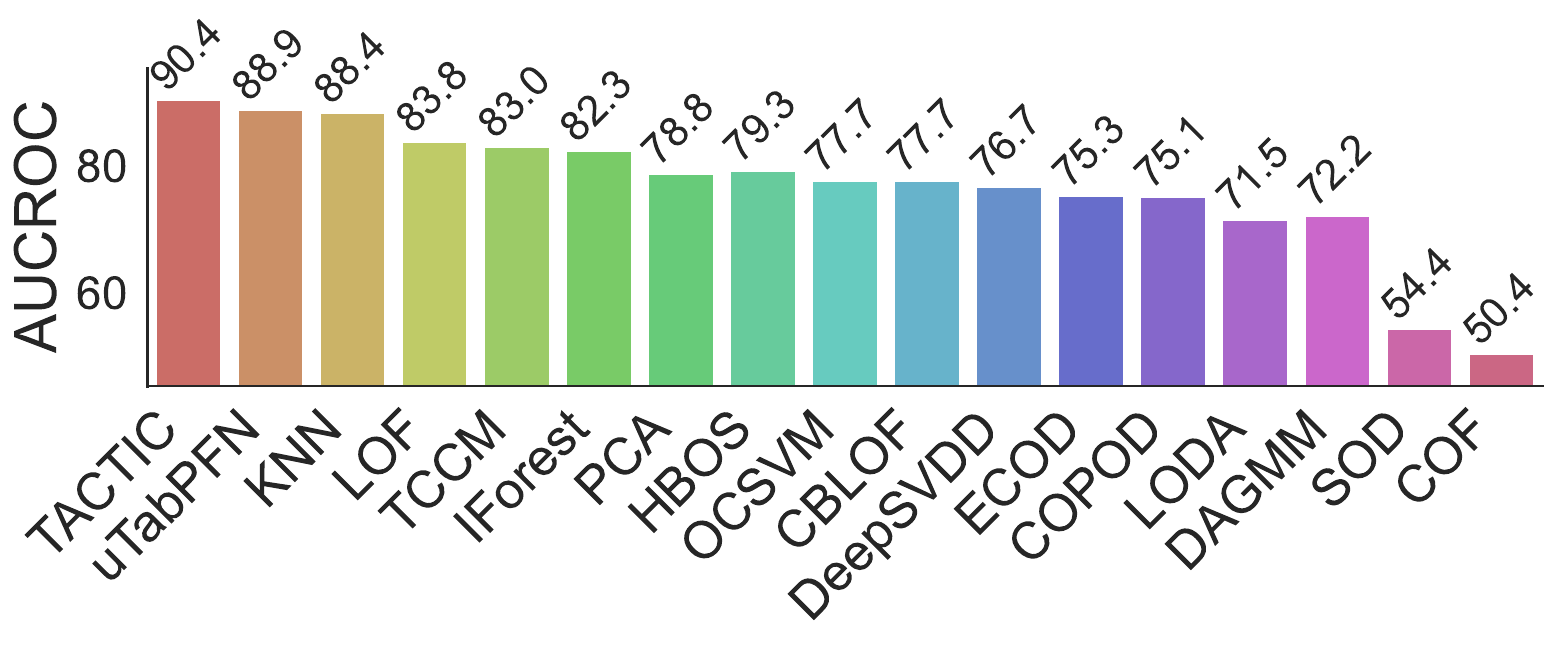}
        \caption{Average AUCROC.}
    \end{subfigure}
    \hfill
    \begin{subfigure}[b]{0.44\textwidth}
        \includegraphics[height=2.5cm]{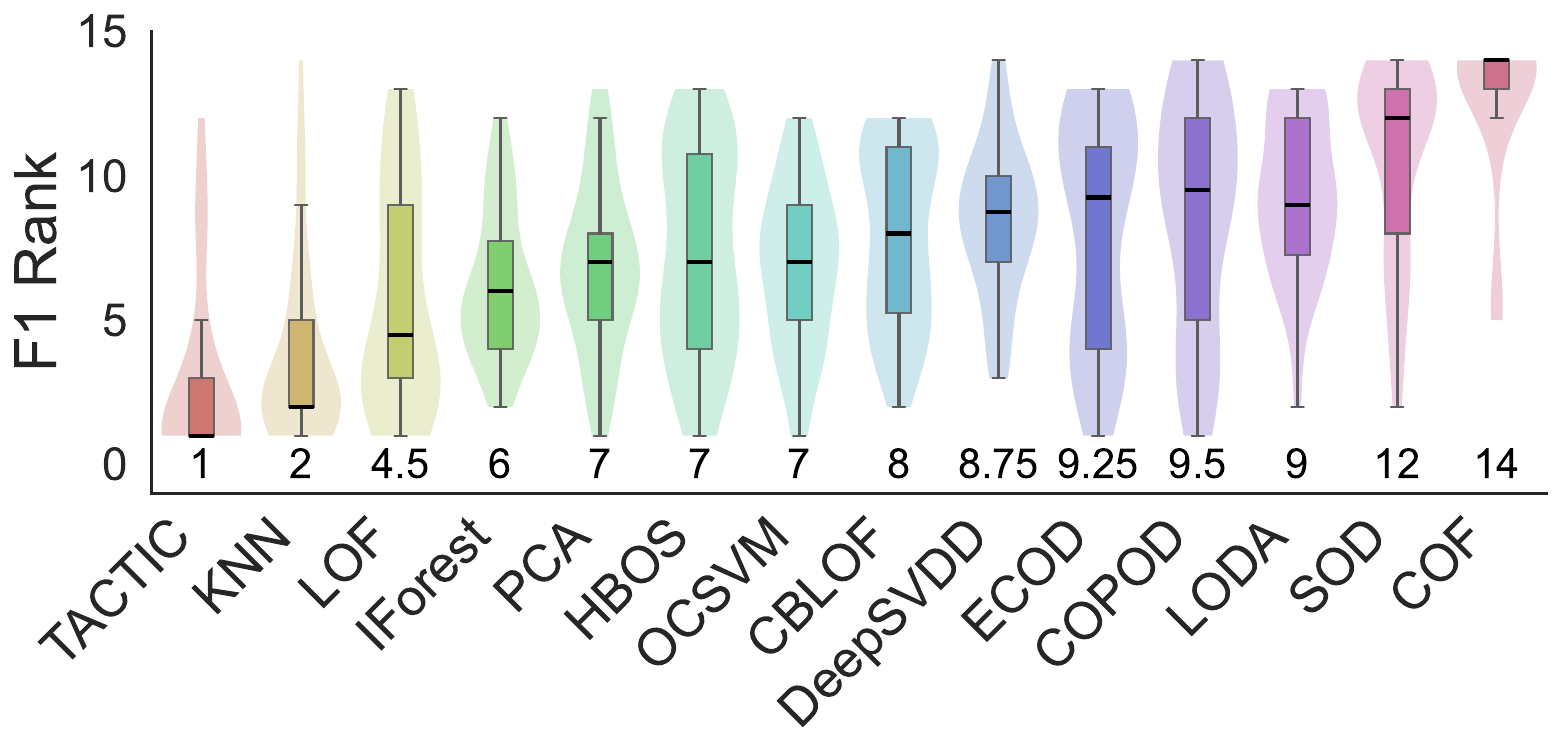}
        \caption{Distribution of F1 ranks.}
    \end{subfigure}
    \hfill
    \begin{subfigure}[b]{0.55\textwidth}
        \includegraphics[height=2.5cm]{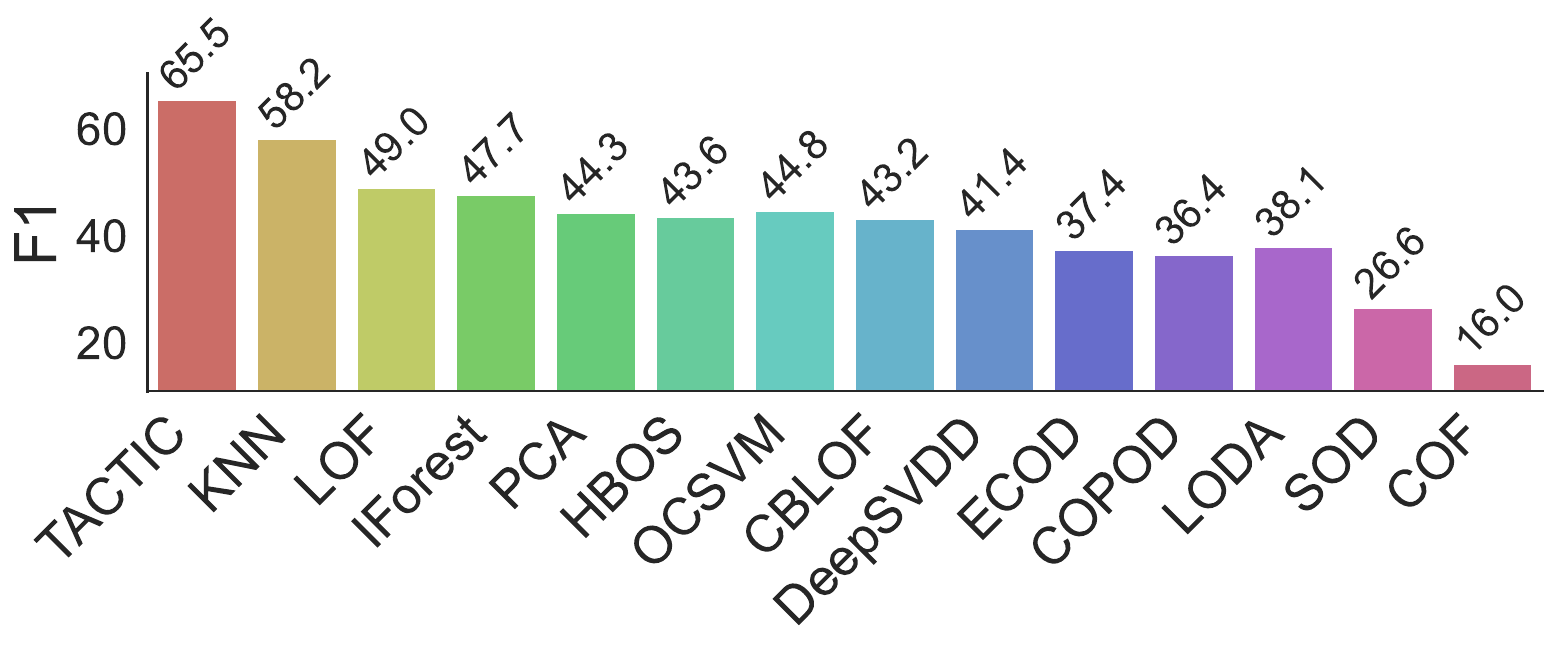}
        \caption{Average F1.}
    \end{subfigure}
    
    \caption{
Performance of our model (\our{}) against classical, deep, and in-context baselines (the unsupervised extension of TabPFN2.5, termed uTabPFN) in the clean-context setting, computed on multiple real-world datasets. The numeric values under the distribution plots indicate the average rank of each method.
    }
    \label{fig:no_anomalies}
\end{figure}

We begin by examining whether an in-context model trained on artificial data can be employed effectively in real-world anomaly detection tasks.
In the baseline setting, we assume an idealized scenario in which the context set is clean and contains only nominal data samples (no unlabeled anomalies). We evaluated our base model, which was pretrained exclusively with clean contexts, against standard anomaly detection benchmarks (fitted on clean training sets as well). 

Figure~\ref{fig:no_anomalies}(a) 
shows that \our{} achieves the best average AUCROC rank, with a narrow rank distribution, indicating both strong performance and consistency across datasets. This ranking advantage is also reflected in the absolute AUCROC values (plot (b)), where our model achieves the highest average AUCROC among all the methods evaluated. %
An even stronger advantage of \our{} is observed for the F1 score.
Figures~\ref{fig:no_anomalies}(c) and~(d) show that \our{} again achieves the best average rank and the highest mean F1, suggesting that the gains are not limited to threshold-independent metrics, but also translate into improved decision quality. 
Overall, these results demonstrate that an in-context model pretrained with anomaly-centric priors can not only generalize to real-world anomaly detection tasks, but also surpass both traditional task-specific methods and general-purpose in-context models.

\subsection{Is In-Context Model Robust to Contaminated Contexts?}
\label{sec:noisy_context}

It is commonly impractical to assume that the training set is free of anomalies. 
In this section, we study how the presence of unlabeled anomalies in the context set affects in-context anomaly detection.
For each dataset, we introduced a controlled number of anomalies in the training set, matching the empirical anomaly rate observed in that dataset.  

\begin{figure}[htpb!]
    \centering
    
    \begin{subfigure}[b]{0.44\textwidth}
        \includegraphics[height=2.5cm]{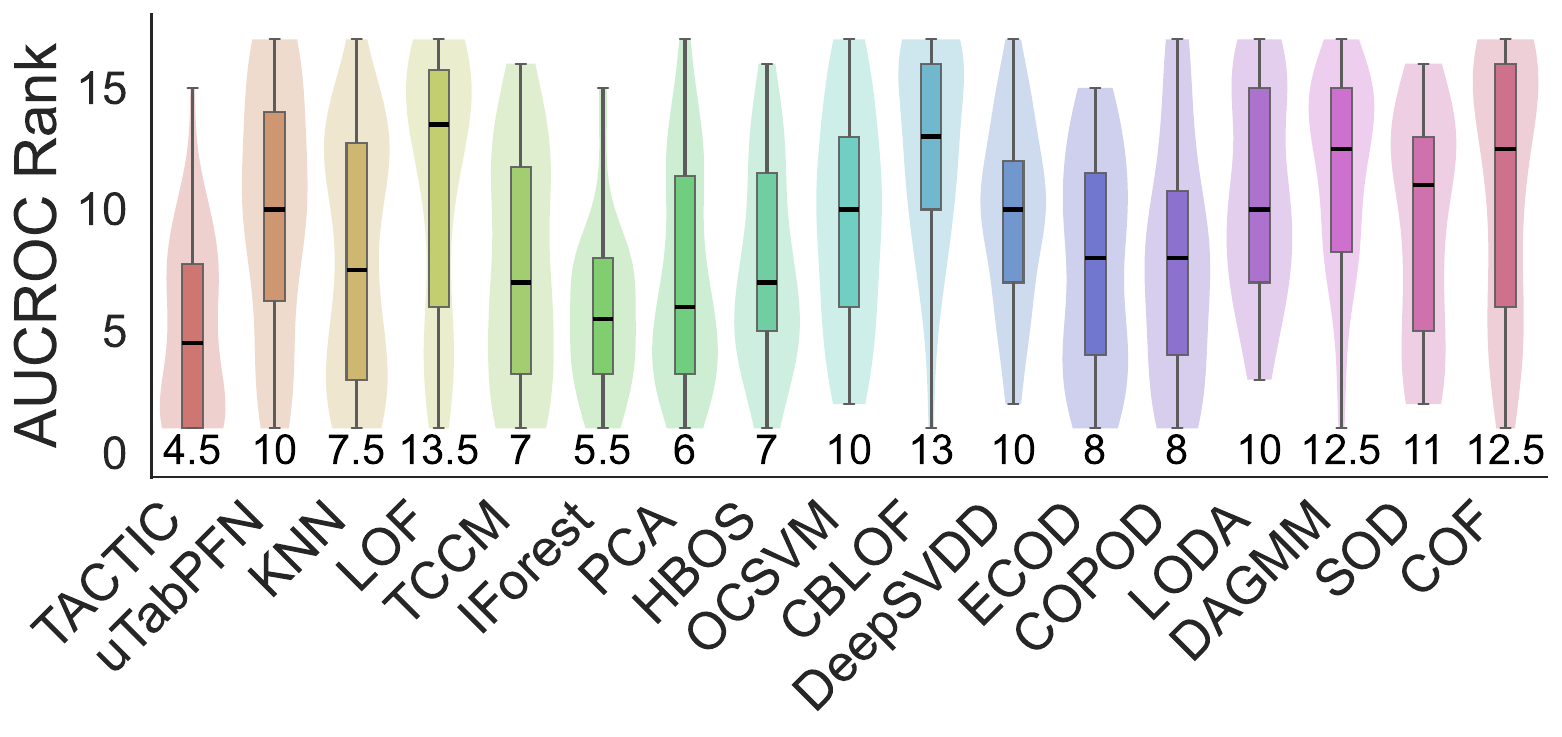}
        \caption{Distribution of AUCROC ranks.}
    \end{subfigure}
    \hfill
    \begin{subfigure}[b]{0.55\textwidth}
        \includegraphics[height=2.5cm]{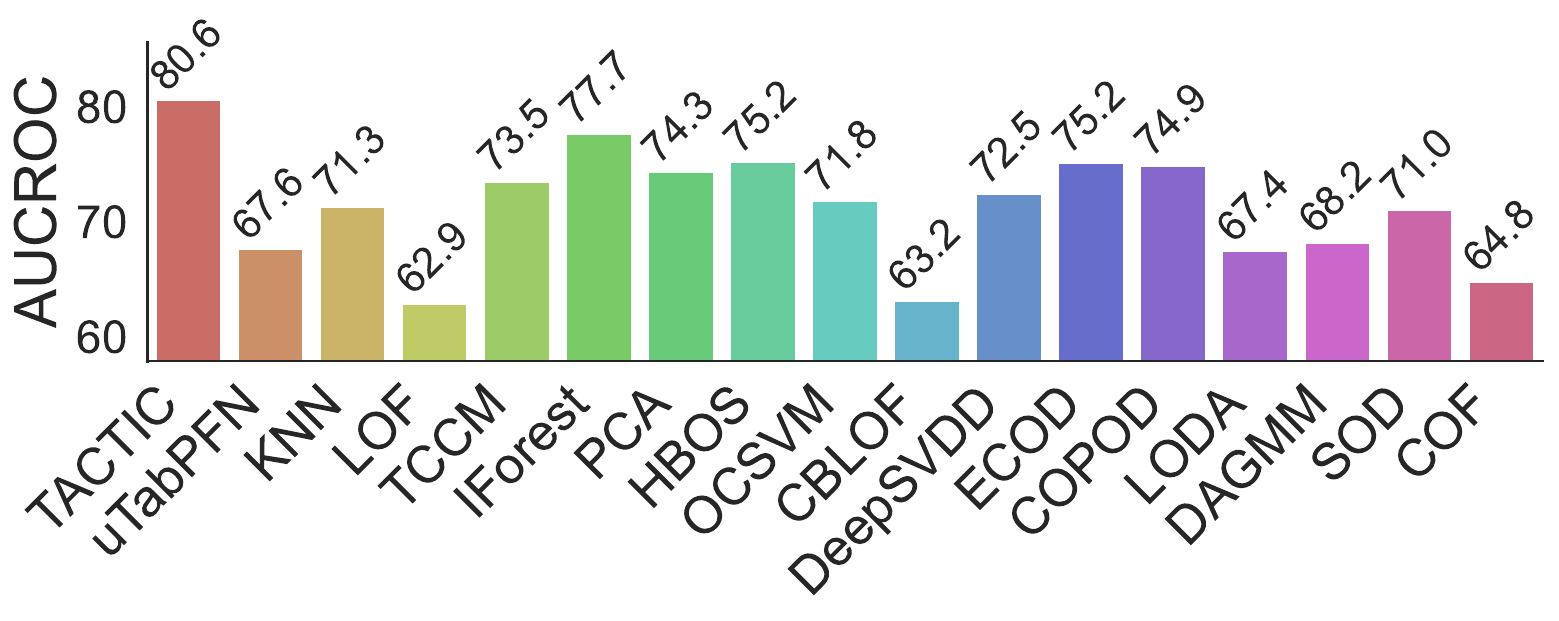}
        \caption{Average AUCROC. }
    \end{subfigure}
    \hfill
    \begin{subfigure}[b]{0.44\textwidth}
        \includegraphics[height=2.5cm]{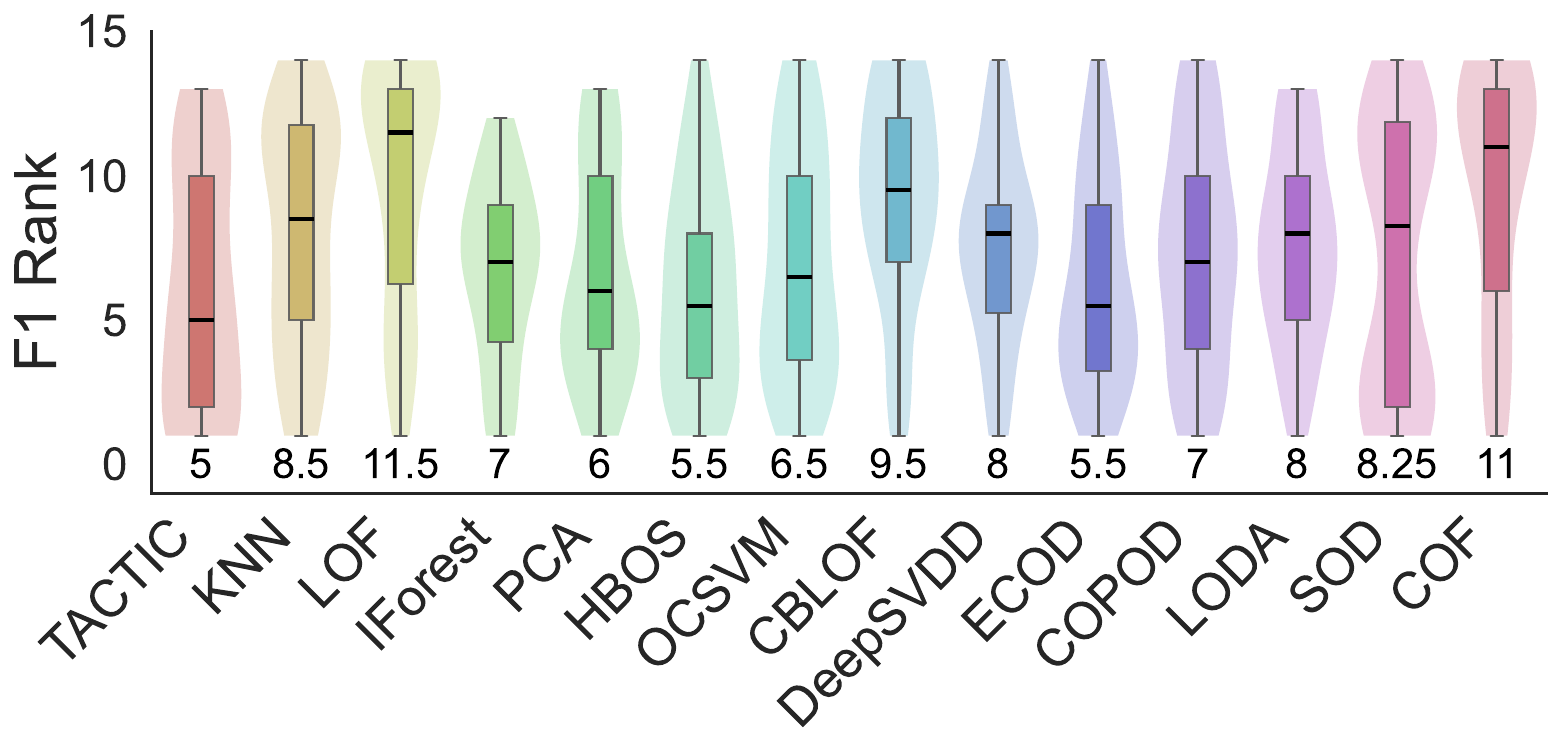}
        \caption{Distribution of F1 ranks.}
    \end{subfigure}
    \hfill
    \begin{subfigure}[b]{0.54\textwidth}
        \includegraphics[height=2.5cm]{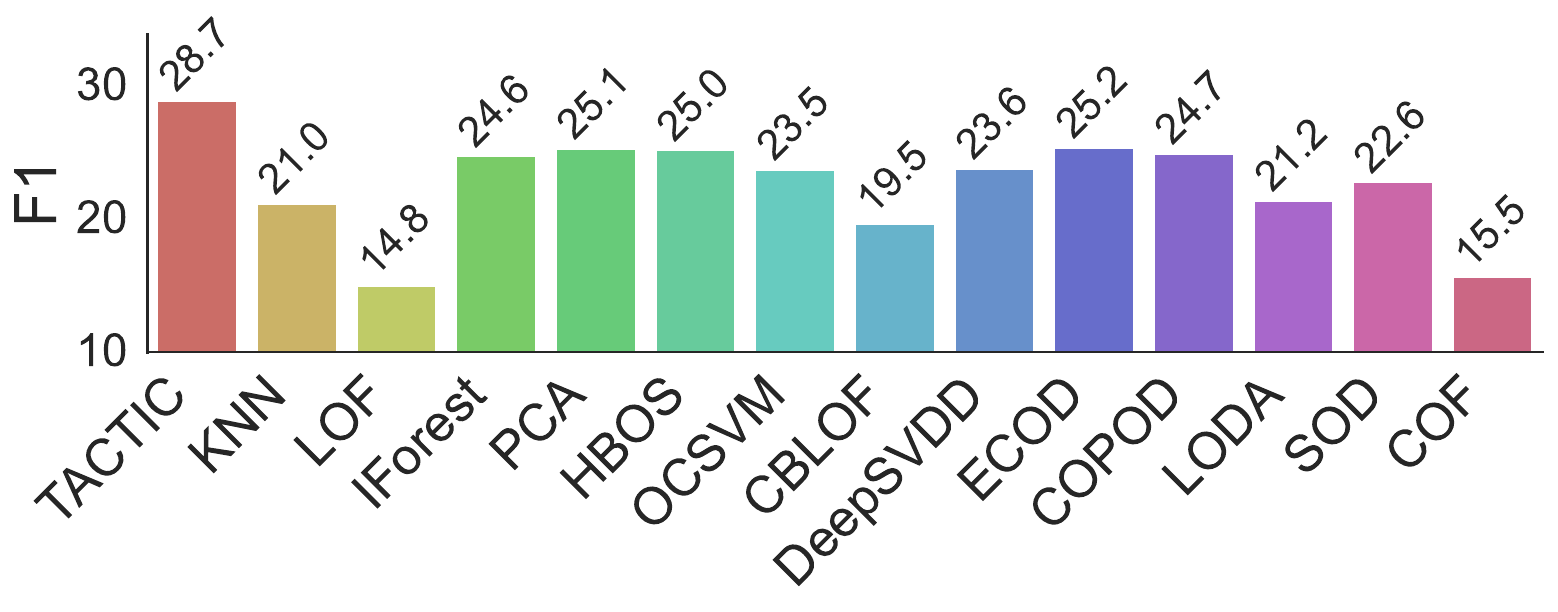}
        \caption{Average F1. }
    \end{subfigure}
    
    \caption{Performance of our model (\our{}) against classical, deep, and in-context baselines  
    in the contaminated (noisy) context setting, computed on multiple real-world datasets. The numeric values under the distribution plots indicate the average rank of each method.
}
\label{fig:contaminated}
\end{figure}

Figure~\ref{fig:contaminated} summarizes the performance of \our{} pretrained on noisy contexts vs. competing methods in the setting with contaminated context. Compared to the clean-context results in Figure~\ref{fig:no_anomalies}, the presence of unlabeled anomalies in the context leads to a noticeable degradation in performance for most baselines, particularly for methods that rely on density estimation. This effect is especially pronounced for uTabPFN2.5, whose average AUCROC rank deteriorates substantially, accompanied by increased variability across datasets (Figure~\ref{fig:contaminated}~(a)). 

In contrast, \our{} pretrained on contaminated contexts,
remains the top-performing method under noisy conditions. As shown in Figures~\ref{fig:contaminated}~(a) and~(b), it achieves the best average AUCROC rank and the highest mean AUCROC among all the approaches evaluated. A similar trend is observed for the F1 score (Figure~\ref{fig:contaminated}~(c,d)), where \our{} again achieves the best average rank and absolute performance. In particular, while absolute scores decrease compared to the clean-context scenario, the relative advantage of \our{} increases, suggesting that pretraining with contaminated contexts enables the model to identify the nominal data distribution even when the context is corrupted. Overall, these results demonstrate that exposing the model to noisy contexts during pretraining is crucial to achieving robust in-context anomaly detection in realistic deployment settings.

\subsection{How Does Context Corruption Affect Performance?}

Building on the previous experiment, we conduct a more systematic analysis of how varying levels of anomalies affect performance. 
In this part, we consider datasets with $k\%$ anomalies in their contexts (queries remained untouched). For each level of anomalies $k \in \{0,5,10,15,20\}$, we selected all datasets containing at least $k\%$ anomalies (note that the number of such datasets available for evaluation decreases as $k$ increases) and adjusted their anomaly rates to match exactly $k\%$. 
For each configuration, we evaluated two variants of our model: one pretrained on clean contexts (TACTIC-C) and one pretrained with noisy contexts (TACTIC-N), and contrasted them with the strongest baselines from previous studies.

    

    


\begin{figure}[thbp]
    \centering
    
    \begin{subfigure}[b]{0.435\textwidth}
        \includegraphics[width=\textwidth]{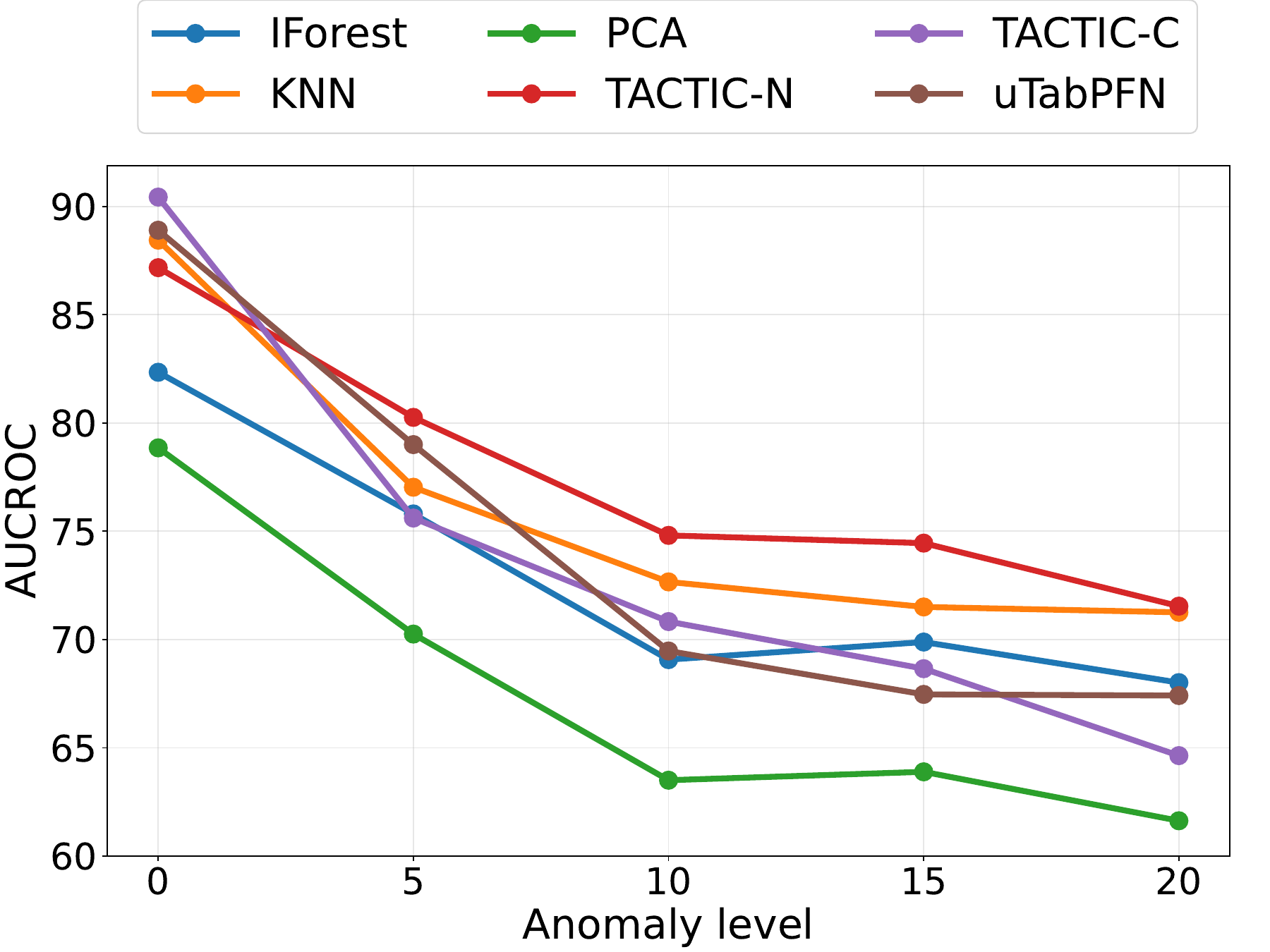}
        \caption{Average AUCROC.}
    \end{subfigure}
    \hfill
    \begin{subfigure}[b]{0.435\textwidth}
        \includegraphics[width=\textwidth]{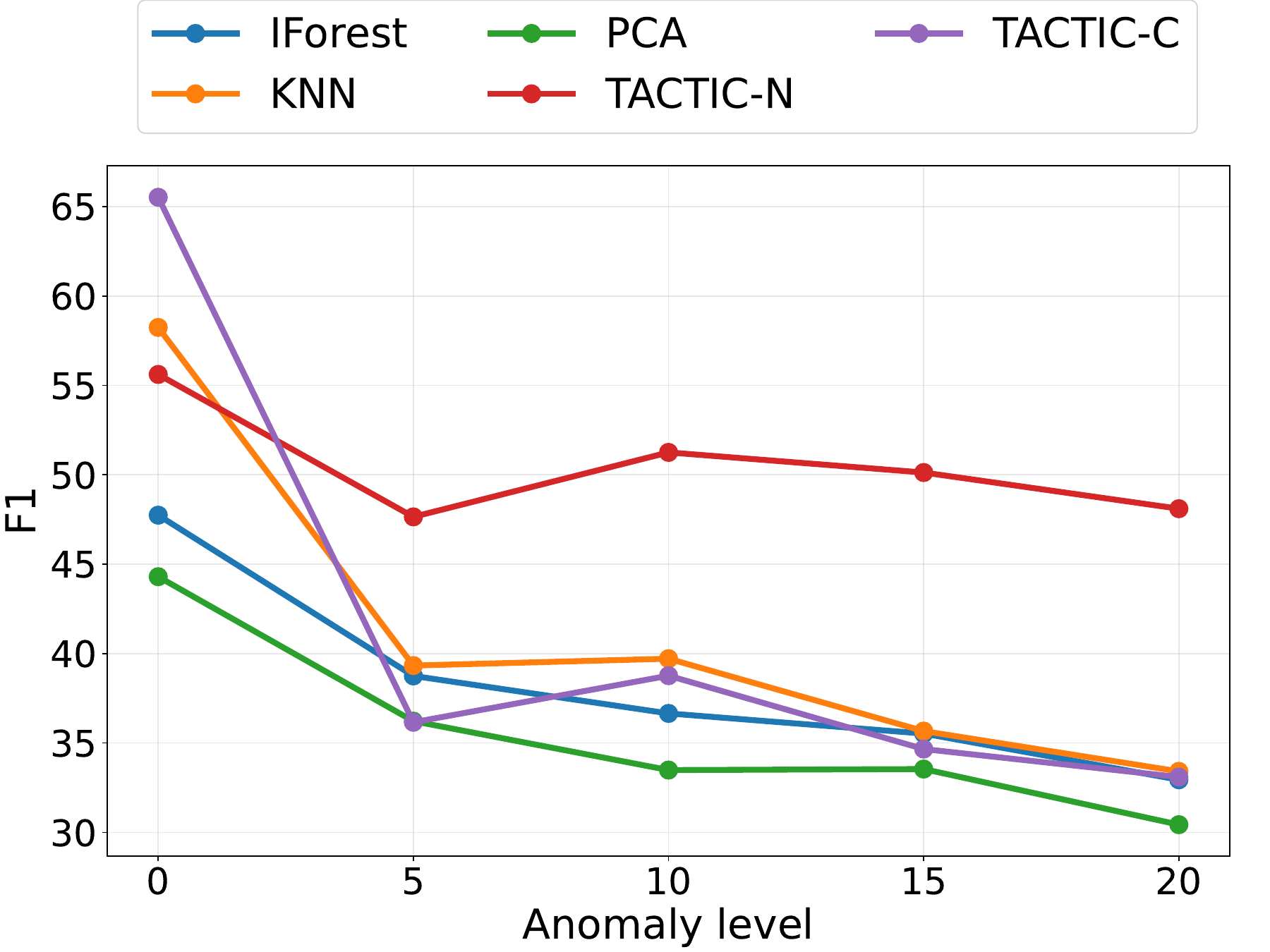}
        \caption{Average F1.}
    \end{subfigure}

    \caption{
Anomaly detection performance averaged over datasets containing at least 
$k\%$ anomalies (note that the number of such datasets decreases as 
k increases),
evaluated using contexts with anomaly levels reduced  to match exactly k\%.
    }
    \label{fig:0_20_anomaly_level}
\end{figure}

Figure~\ref{fig:0_20_anomaly_level} shows that performance decreases for all methods as the noise level increases, but the rate of degradation varies substantially. 
TACTIC-C achieves the highest AUCROC and F1 at $k=0\%$, but context corruption leads to noticeable drops in its performance, and it eventually approaches the performance of the classical baselines. 
In contrast, TACTIC-N achieves top or near-top performance and remains the most robust across all noise levels.
For F1, its dominance becomes even more pronounced (a slight increase in F1 from $k=5\%$ to $k=10\%$ can be attributed to the shrinking pool of datasets at larger $k$).
Among the baselines, IForest and KNN remain competitive under small noise levels, but exhibit smooth degradation. Overall, they never match TACTIC-C in clean contexts nor the robustness of TACTIC-N. PCA consistently underperforms. Crucially, uTabPFN2.5 shows a high sensitivity to context contamination. Although it is competitive in the clean setting (see Section~\ref{sec:clean_context}), its AUCROC decreases rapidly as noise increases. This behavior aligns with the discussion in Section~\ref{sec:noisy_context}, where likelihood-based scoring induced by a classification-oriented prior was observed to be unstable under distribution shift. Without explicit robustness mechanisms, uTabPFN2.5 struggles with contaminated contexts.


Overall, our results reinforce the conclusions from Sections~\ref{sec:clean_context} and~\ref{sec:noisy_context}: clean-context pretraining yields strong performance under ideal conditions, but robustness depends strongly on the pretraining distribution and can be achieved by pretraining on corrupted contexts. Exposure to noisy contexts during training improves robustness to corruption, while models based on fixed density-based scoring, such as uTabPFN2.5, remain sensitive to noisy contexts.




\subsection{Can One Model Handle Multiple Anomaly Types?}

Real anomalies are heterogeneous and often arise from different processes. To study their varying impacts, we construct synthetic datasets augmented with three qualitatively distinct types of anomalies and test whether a single in-context model can robustly handle diverse anomalies. 
In particular, we constructed synthetic Gaussian mixture datasets augmented with 
\emph{local}, \emph{global}, and \emph{cluster} anomalies (see Section~\ref{sec:priors} for detail) following ADBench design~\cite{han2022adbench}. For each test run, anomaly rates were sampled uniformly from 5-25\% for both the context and query sets and we generated 25 test sets for each anomaly type. The baselines include the strongest methods from the ADBench benchmark, together with TCCM and the in-context baseline uTabPFN2.5. Our model was evaluated in the variant (N) pretrained with contaminated contexts.

\begin{figure}[ht!]
    \centering
    
    \begin{subfigure}[b]{0.9\textwidth}
\includegraphics[width=0.49\textwidth]{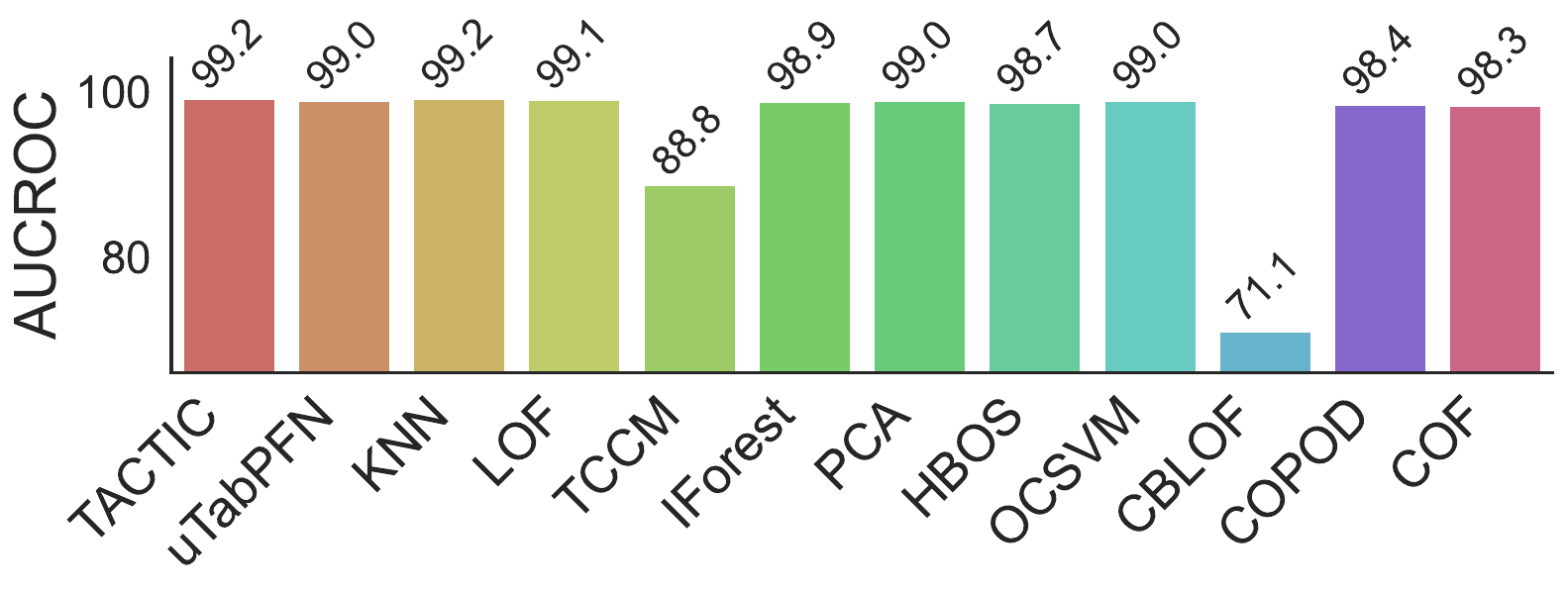}
\includegraphics[width=0.49\textwidth]{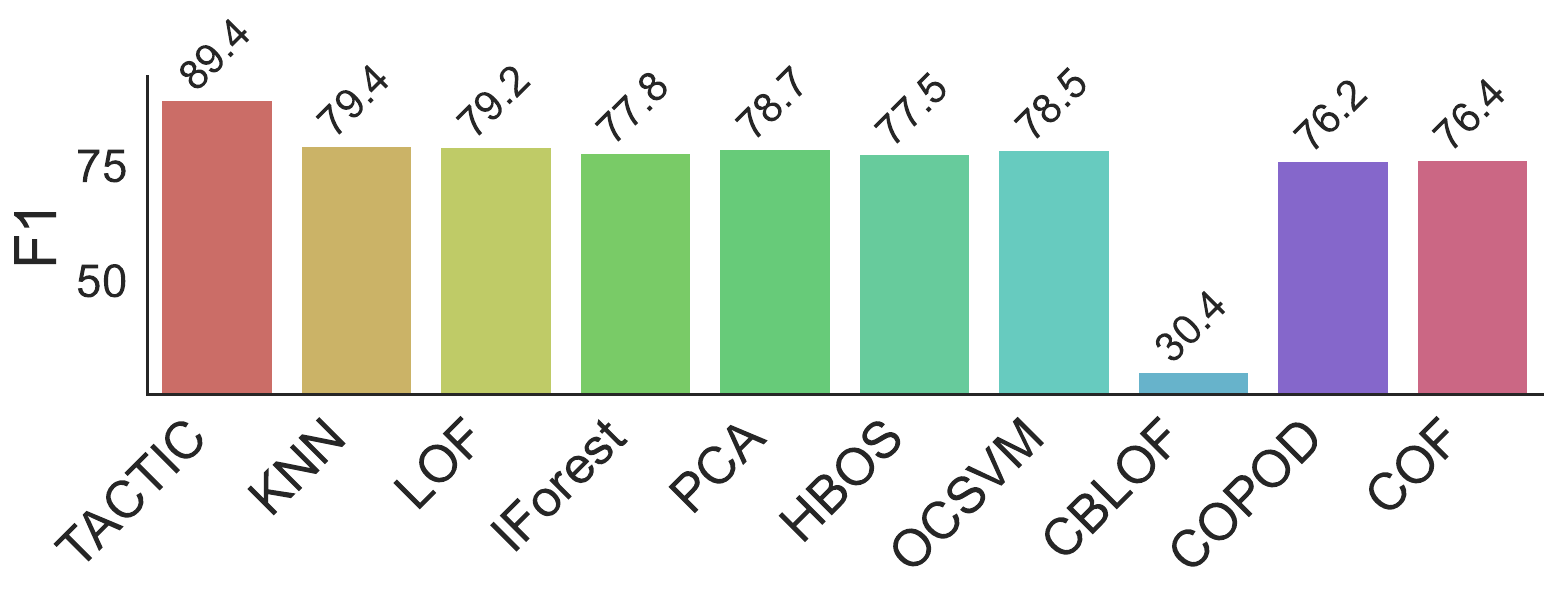}
        \caption{Local anomalies}
    \end{subfigure}
    \hfill
    \begin{subfigure}[b]{0.9\textwidth}
      \includegraphics[width=0.49\textwidth]{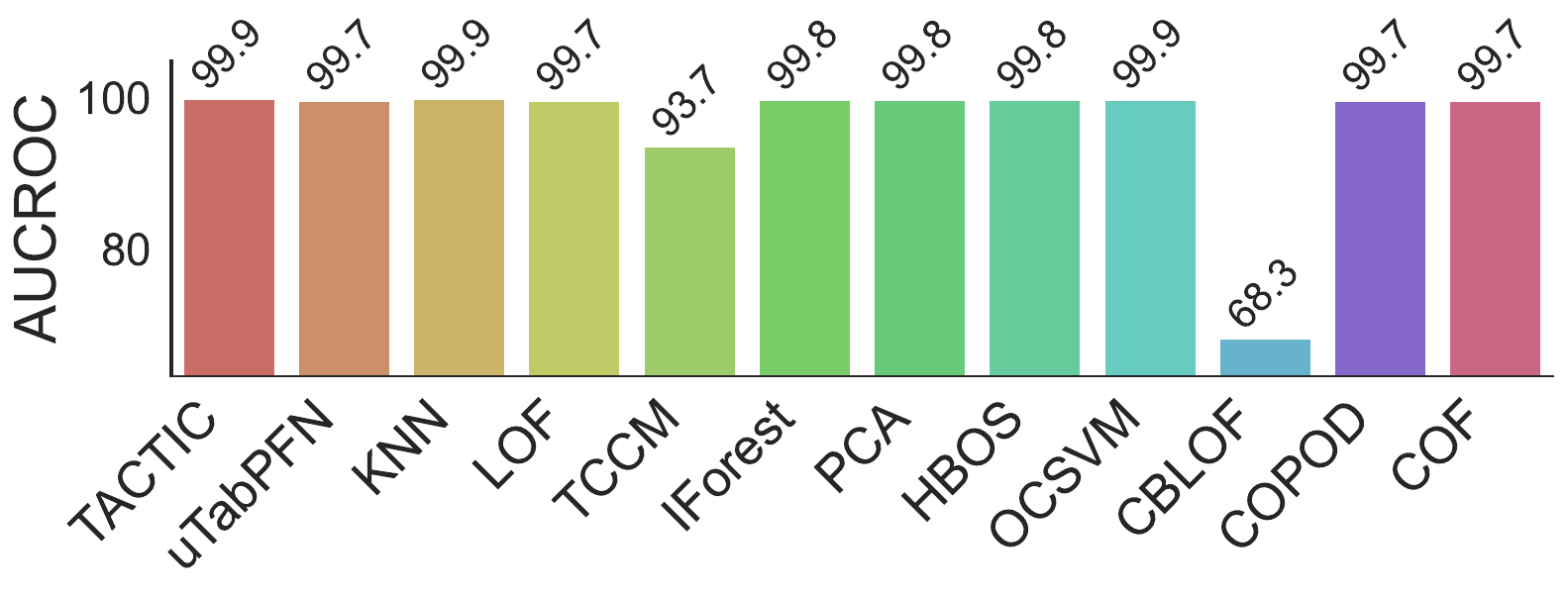}
      \includegraphics[width=0.49\textwidth]{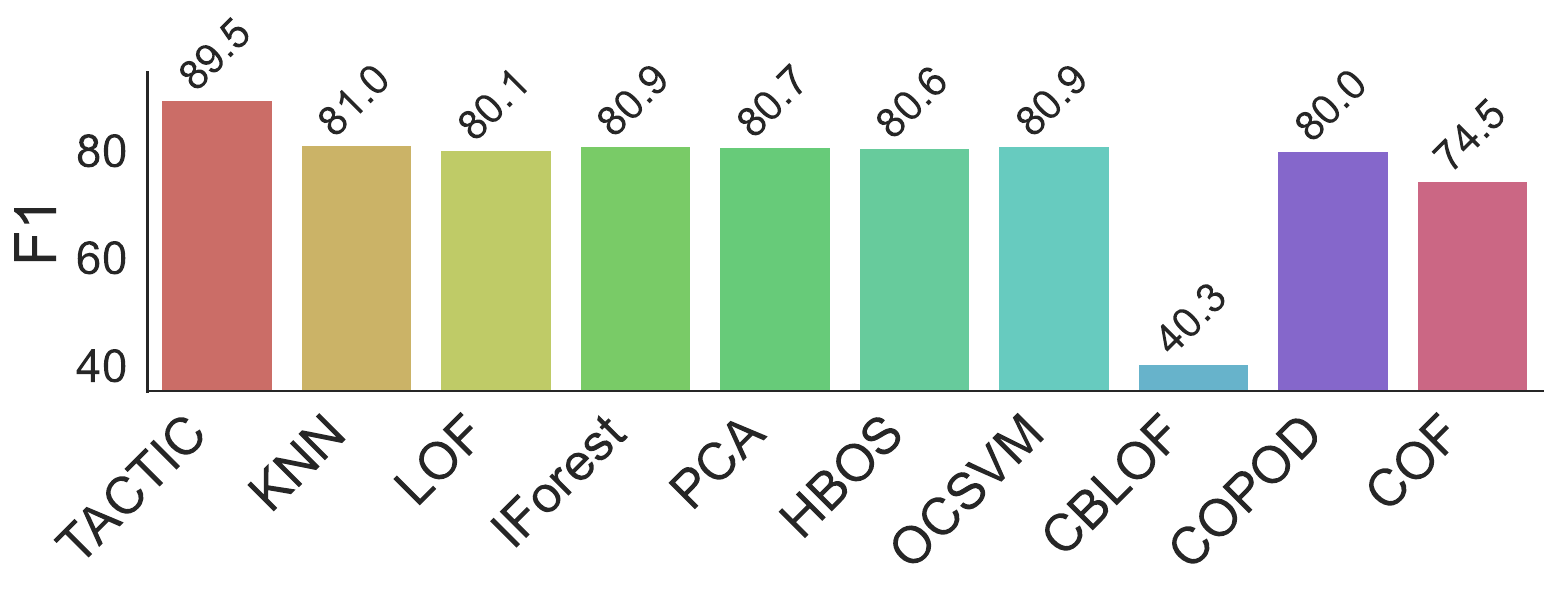}
        \caption{Global anomalies}
    \end{subfigure}
    \hfill
    \begin{subfigure}[b]{0.9\textwidth}
        \includegraphics[width=0.49\textwidth]{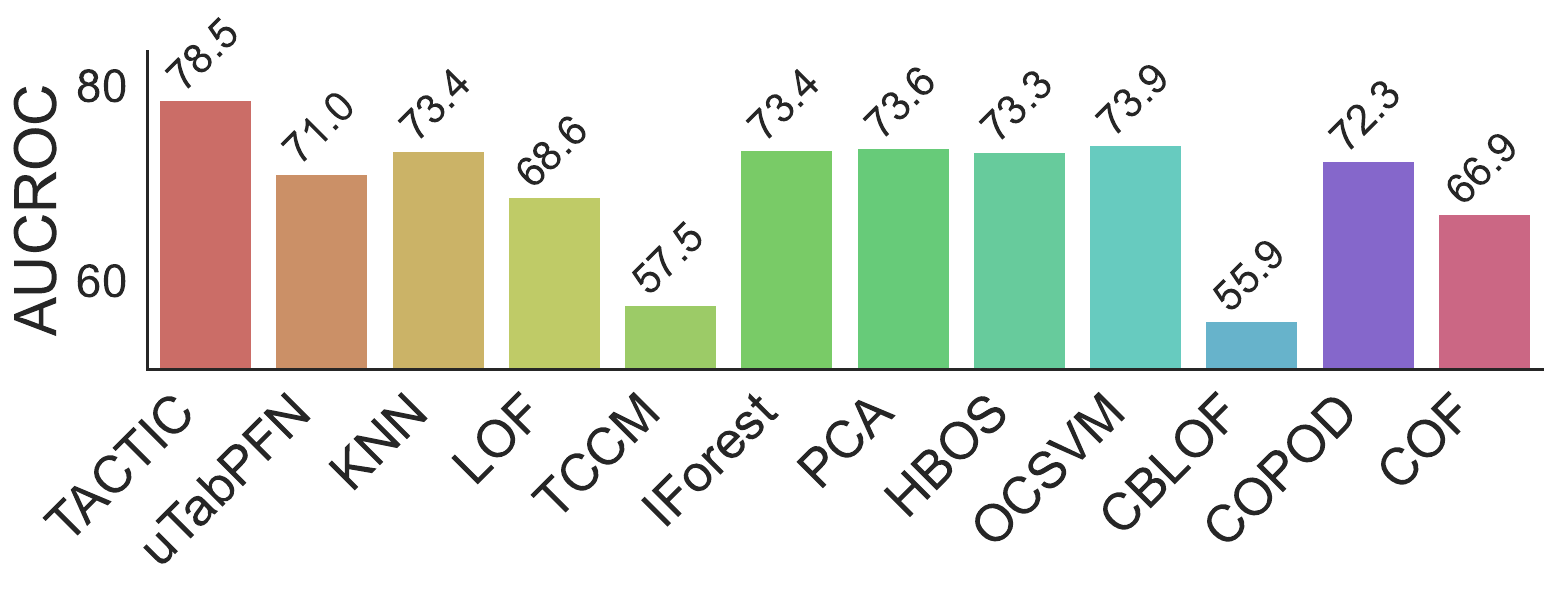}
        \includegraphics[width=0.49\textwidth]{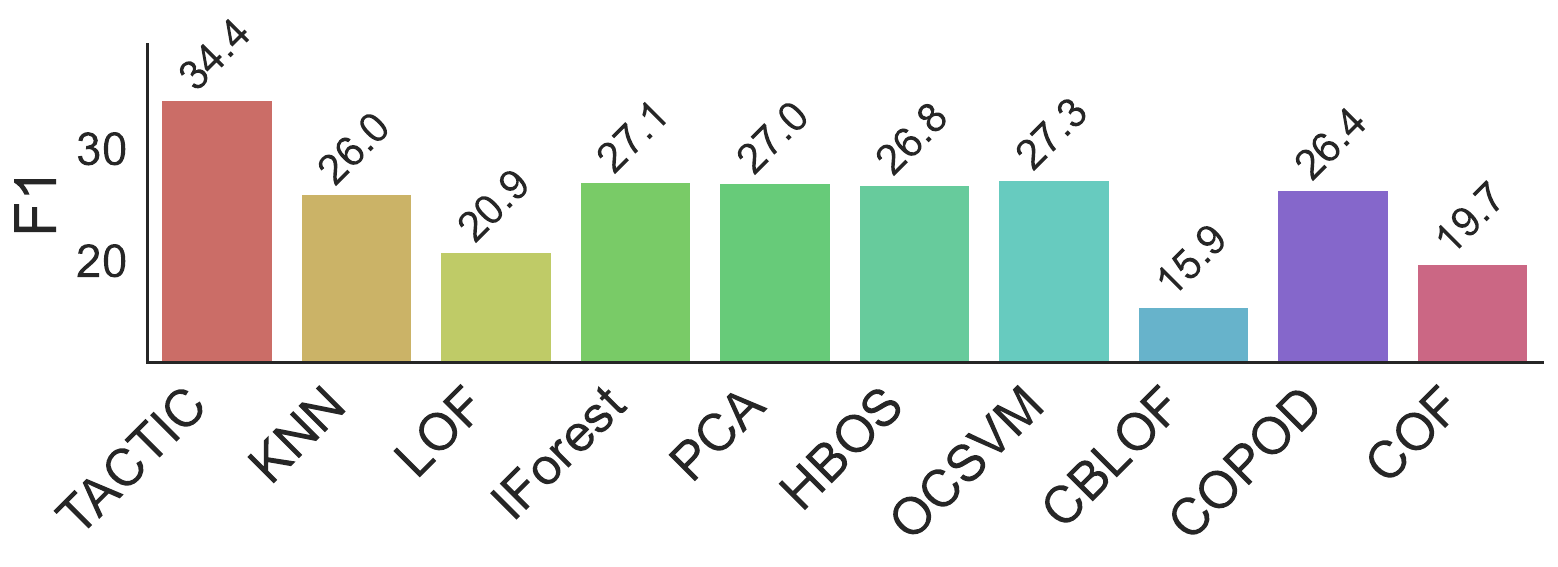}
        \caption{Cluster anomalies}
    \end{subfigure}
    
    \caption{Performance on synthetic Gaussian mixture datasets with different anomaly types. Anomalies were added to both contexts and query sets.}
    \label{fig:gmm_anomalies}
\end{figure}

It is evident from Figure~\ref{fig:gmm_anomalies} that for local anomalies, which are close to high-density regions, 
TACTIC achieves both the highest AUCROC and F1. 
Although 
many competing methods reach similar AUCROC values,  their F1 scores are noticeably lower. 
The weaker performance of the baselines points toward their poorer calibration and worse precision-recall trade-offs. Similar characteristics
are observed  
for global anomalies, which lie far from the support of nominal data.
Most methods reaches nearly perfect AUCROC, yet only TACTIC maintains an impressive F1 score. 
Finally, for cluster anomalies, which form shifted groups, anomaly detection is the most difficult. The performance of many methods drops substantially. For example, uTabPFN2.5 degrades noticeably, reaching an AUCROC of only $71\%$. Although TACTIC also shows lower performance for cluster anomalies, it remains the strongest performer in both AUCROC and F1. 

These findings indicate that the proposed model generalizes well across fundamentally different anomaly types. While previous sections showed strong performance on real datasets and robustness to noisy contexts, this experiment confirms that the same pretrained model can handle diverse anomaly types without additional specialization. 


\subsection{Which Priors Matter for Pretraining?}

\begin{figure}[t!]
    \centering
    
    \begin{subfigure}[b]{0.45\textwidth}
        \includegraphics[width=\textwidth]{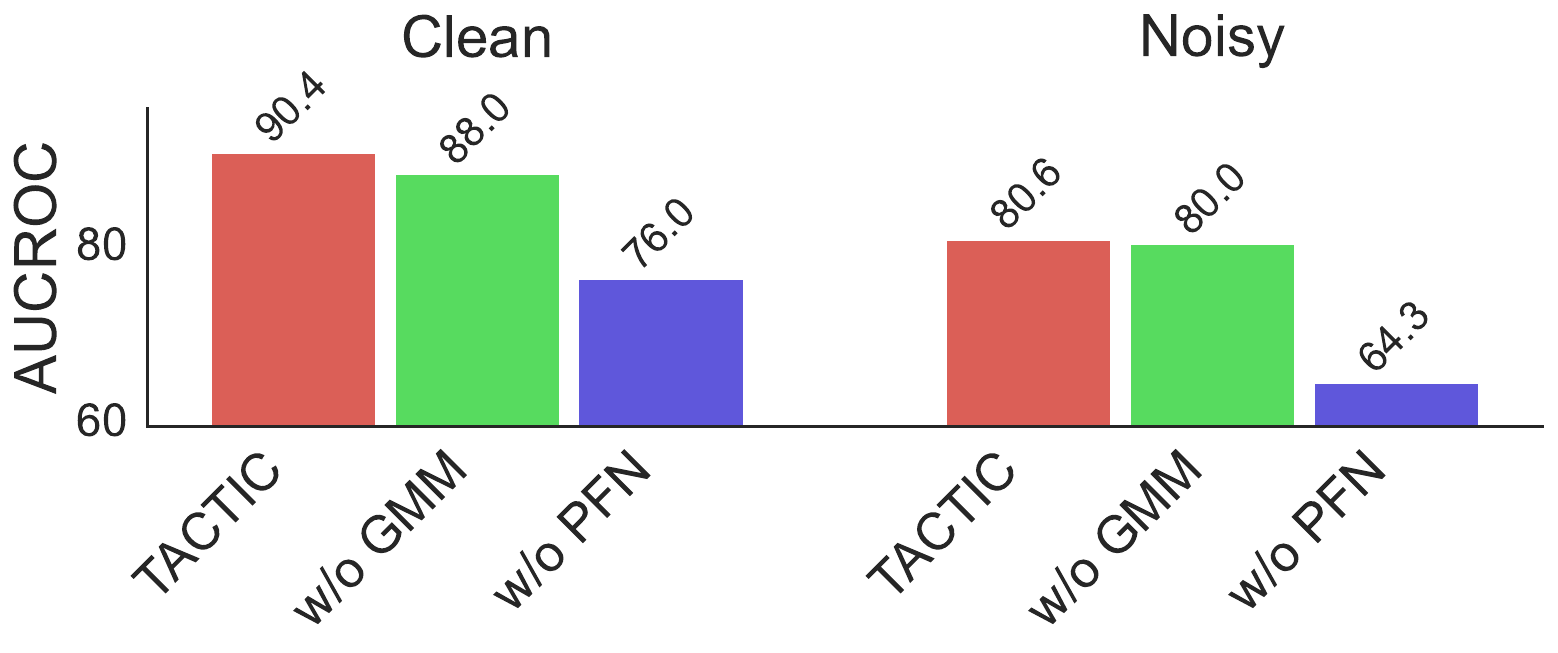}
        \caption{Average AUCROC.}
    \end{subfigure}
    \hfill
    \begin{subfigure}[b]{0.45\textwidth}
        \includegraphics[width=\textwidth]{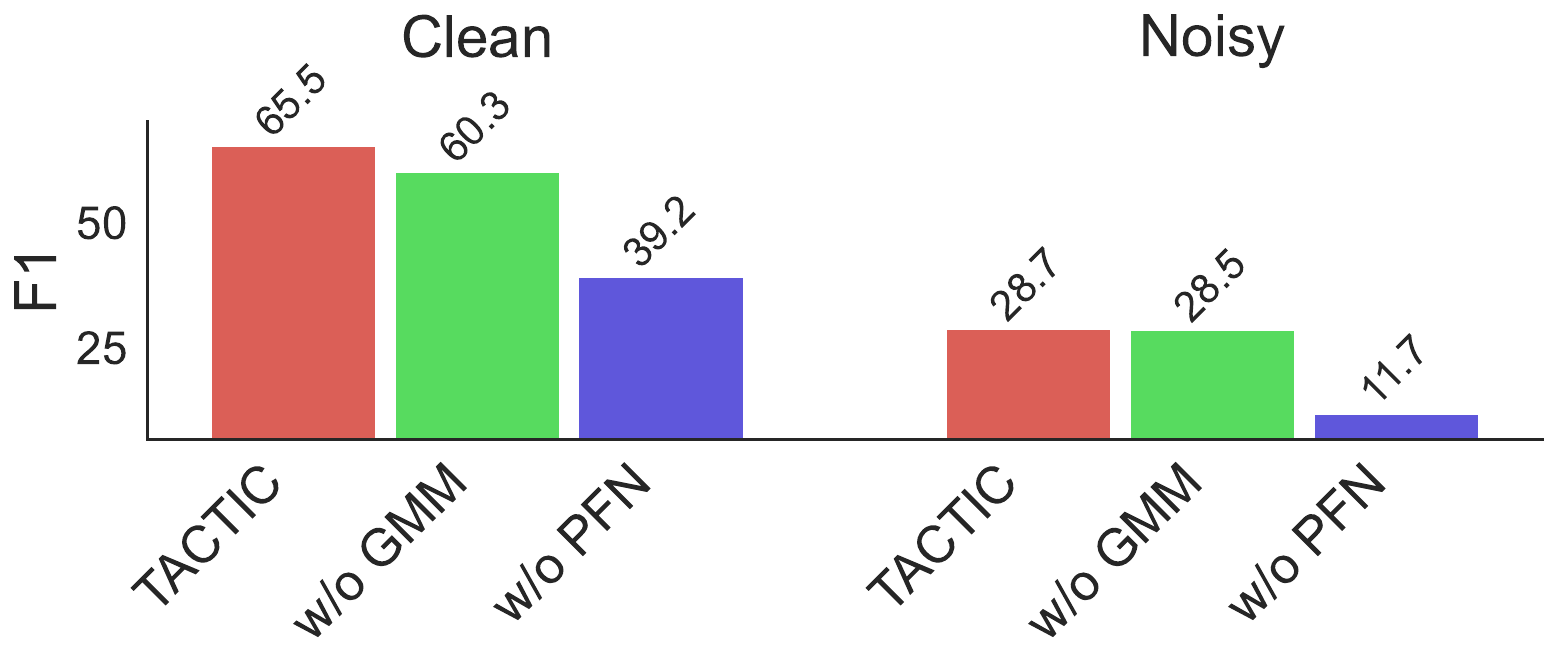}
        \caption{Average F1.}
    \end{subfigure}

    \caption{Impact of priors used for pretraining on performance. We compare the default TACTIC models pretrained on clean or contaminated (noisy) contexts against variants with GMM-based priors excluded (w/o GMM) and without classification-based priors (w/o PFN). }
    \label{fig:priors_ablation}
\end{figure}

\our{} was pretrained on a mixture of two prior families: (i) classification-based PFN-style priors, and (ii) anomalies generated sythetically for GMM-based prior. To assess their respective contributions on model quality and robustness, we perform an ablation study in which we remove one prior component at a time and evaluate the resulting models in both clean and noisy context settings (see Figure~\ref{fig:priors_ablation}).

In the clean setting, removing the GMM-based priors leads to a moderate decrease (AUCROC drops from $90.4\%$ down to $88.0\%$  and F1 from $65.5\%$ down to $60.3\%$). On the other hand, removing the PFN-style prior causes a more substantial drop (respectively down to $76.0\%$ and  $39.2\%$). In the noisy setting, the same pattern persists. The full model remains the strongest (AUCROC=$80.6\%$, F1=$28.7\%$). Excluding GMM priors slightly reduces performance (down to $80.0\%$ and $28.5\%$), while excluding the PFN-style prior results in a large degradation (down to $64.3\%$ and $11.7\%$). 

We empirically conclude that, the classification-based (PFN-style) prior has a stronger impact on discriminative performance and robustness to noise, and removing it leads to a larger degradation for both clean and contaminated settings. However, the GMM-based component is also necessary for full generalization. Excluding either subprior reduces performance, indicating that both are crucial.
These findings align with the theoretical motivation from Section~\ref{sec:priors}, as the mixture prior, combining the PFN-style and GMM-based components, introduces complementary biases across anomaly types and latent structures.




\section{Related Work}

\paragraph{Anomaly detection:} Existing approaches to unsupervised anomaly detection can be broadly divided into classical (shallow) methods and deep learning-based models. Classical parametric techniques such as Gaussian Mixture Models (GMMs)~\cite{agarwal2007detecting} detect anomalies via low likelihood under an assumed distribution, while robust covariance estimators and subspace methods like PCA~\cite{pca} rely on reconstruction or projection errors. Nonparametric density estimators~\cite{hbos} relax distributional assumptions but scale poorly in high dimensions. Distance- and neighborhood-based methods, including kNN~\cite{knn}, LOF~\cite{lof}, and CBLOF~\cite{cblof}, measure abnormality relative to the local data structure. Isolation Forest~\cite{isoforest} isolates anomalies through random partitioning and is generally more scalable than density-based methods. The One-Class SVM (OCSVM)~\cite{ocsvm}  represents a discriminative approach that learns a decision boundary around normal data in the feature space. Although simple and interpretable, classical methods struggle with complex nonlinear patterns, high dimensionality, and large-scale datasets.

Deep models address these limitations by learning representations tailored for anomaly detection. The Two-stage approaches first learn latent features using autoencoders (AEs) or VAEs, then apply a separate anomaly detector~\cite{an2015variational}. 
End-to-end methods, such as DeepSVDD~\cite{deepsvdd}, jointly optimize representation learning and one-class objectives, but often require architectural constraints to avoid collapse. GAN-based methods (e.g., GANomaly~\cite{akcay2018ganomaly}) model normal data distributions but suffer from adversarial instability. Hybrid density-based models like DAGMM~\cite{zong2018deep} combine reconstruction with latent distribution modeling, increasing architectural complexity. Recent advancements were made by modeling data distribution using diffusion models~\cite{livernoche2023diffusion} or predicting a time-conditioned contraction vector toward a fixed target using flow matching~\cite{tccm}. 
Most anomaly detection methods estimate anomaly scores rather than discriminate nominal data from anomalies, which causes problems in their practical usage. OneFlow~\cite{maziarka2021oneflow} turns a generative flow model into a discriminative approach following the idea of OCSVM~\cite{ocsvm}.

\paragraph{In-Context Learning:}

TabPFN~\cite{tabpfn} has been the first tabular foundation model that inspired other works in this area~\cite{den2024fine,tabicl,zhangmitra}. Originally introduced for classification, several extensions\footnote{\url{https://docs.priorlabs.ai/capabilities/anomaly-detection}} were designed to apply it for unsupervised tasks such as density estimation, anomaly detection, or clustering without additional fine-tuning~\cite{hollmann2025accurate}. In addition to the general TabPFN repository, ZEUS~\cite{zeus} designed a clustering-centric prior for pretraining a zero-shot model for clustering tabular data, which was later developed in~\cite{zhao2026tabclustpfn}. The authors of \cite{ma2023tabpfgen,margeloiu2024tabebm} devised energy-based generative models, in which a pretrained TabPFN is used to define a class-conditional energy.


\section{Conclusion}

In this work, we investigated in-context learning as a paradigm for unsupervised anomaly detection in tabular data. Our analysis revealed that directly extending classification-oriented foundation models to anomaly detection leads to sensitivity to noise in contexts, 
in addition 
to high computational cost.
To address these limitations, we introduced \our{}, a discriminative in-context model pretrained on anomaly-centric priors and explicitly designed for robust anomaly detection. Unlike score-based approaches that require post-hoc thresholding, \our{} produces calibrated anomaly probabilities in a single forward pass, enabling fast and unambiguous decisions without dataset-specific tuning. Extensive experiments on real-world and synthetic benchmarks demonstrate that our model consistently outperforms classical, deep, and general-purpose in-context baselines in both clean and noisy settings.  

\section*{Acknowledgments}
This research is part of the project No. \textbf{2022/45/P/ST6/02969} co-funded by the National
Science Centre and the European Union Framework Programme for Research and
Innovation Horizon 2020 under the Marie Skłodowska-Curie grant agreement No.
945339. For the purpose of Open Access, the authors have applied a CC-BY public copyright licence to any Author Accepted Manuscript (AAM) version arising from this submission. 
\\
\includegraphics[width=1cm]{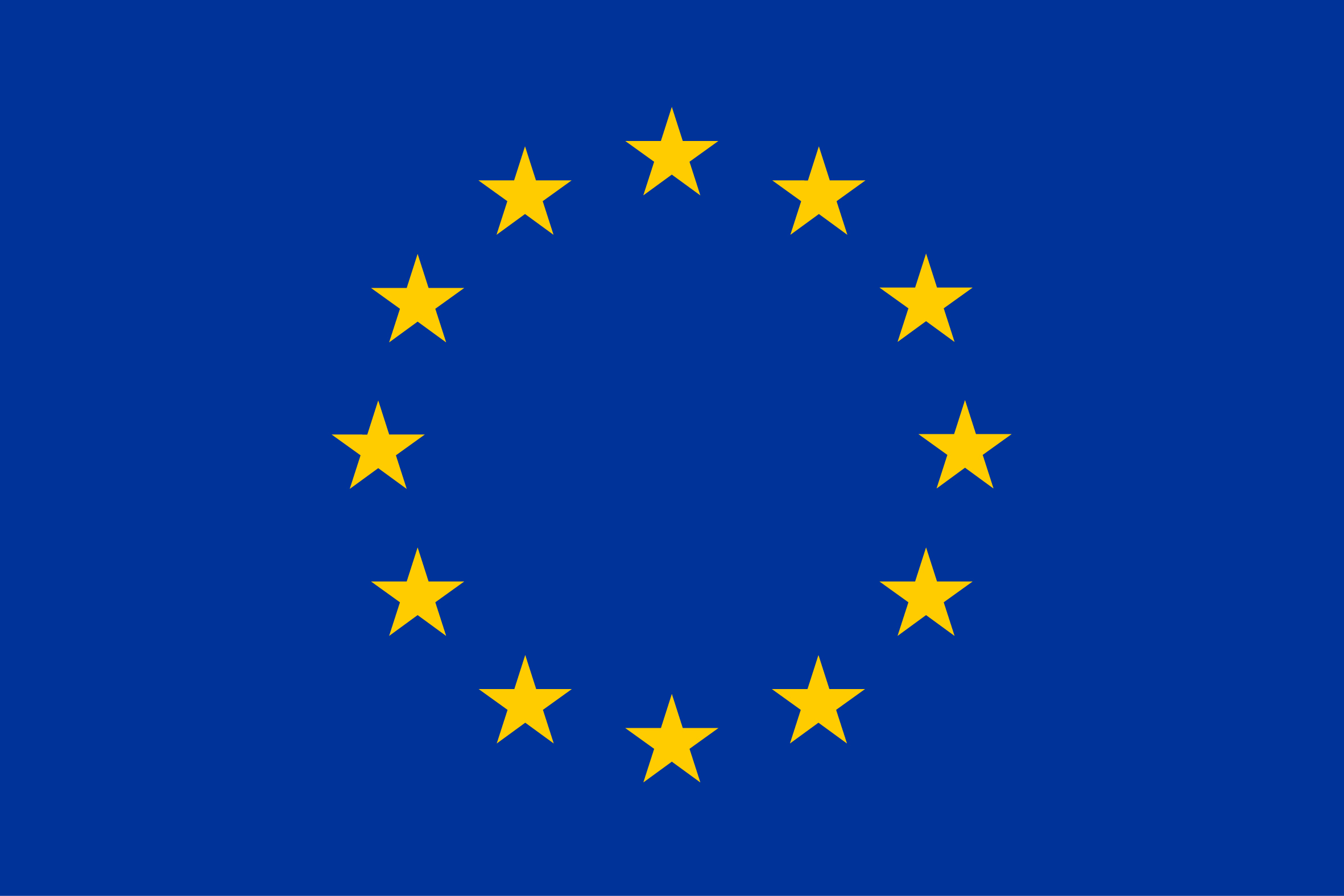} \includegraphics[width=1.9cm]{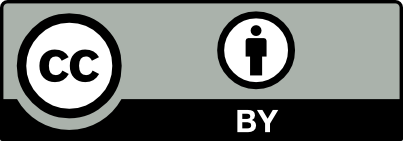}
\\ 
The research of P. Marszałek and M. Śmieja was supported by the National Science Centre (Poland), grant no. \textbf{2023/50/E/ST6/00169}. We also gratefully acknowledge Polish high-performance computing infrastructure PLGrid (HPC Center: ACK Cyfronet AGH) for providing computer facilities and support within computational grant no. \textbf{PLG/2025/018969}.



\clearpage
\appendix

\section{Statistics of real-world datasets}

Table~\ref{app:statics_real} presents summary statistics for the real-world datasets used in our study. The table includes the names of the datasets with their corresponding indexes, the number of samples (\#Samples), the number of features (\#Features), as well as the number of anomalies (\#Anomalies) and their proportion (\%Anomalies) in each dataset.

\begin{table}[!h]
\centering
\scriptsize
\caption{ADbench dataset statistics.}
\label{app:statics_real}
\begin{tabular}{lccccc}
\toprule
Id & Name & \#Samples & \#Features & \#Anomalies & \%Anomalies \\
\midrule
1 & ALOI & 49534 & 27 & 1508 & 3.04 \\
2 & annthyroid & 7200 & 6 & 534 & 7.42 \\
4 & breastw & 683 & 9 & 239 & 34.99 \\
6 & cardio & 1831 & 21 & 176 & 9.61 \\
7 & Cardiotocography & 2114 & 21 & 466 & 22.04 \\
8 & celeba & 202599 & 39 & 4547 & 2.24 \\
10 & cover & 286048 & 10 & 2747 & 0.96 \\
11 & donors & 619326 & 10 & 36710 & 5.93 \\
12 & fault & 1941 & 27 & 673 & 34.67 \\
13 & fraud & 284807 & 29 & 492 & 0.17 \\
14 & glass & 214 & 7 & 9 & 4.21 \\
15 & Hepatitis & 80 & 19 & 13 & 16.25 \\
16 & http & 567498 & 3 & 2211 & 0.39 \\
18 & Ionosphere & 351 & 32 & 126 & 35.90 \\
19 & landsat & 6435 & 36 & 1333 & 20.71 \\
20 & letter & 1600 & 32 & 100 & 6.25 \\
21 & Lymphography & 148 & 18 & 6 & 4.05 \\
22 & magic.gamma & 19020 & 10 & 6688 & 35.16 \\
23 & mammography & 11183 & 6 & 260 & 2.32 \\
27 & PageBlocks & 5393 & 10 & 510 & 9.46 \\
28 & pendigits & 6870 & 16 & 156 & 2.27 \\
29 & Pima & 768 & 8 & 268 & 34.90 \\
30 & satellite & 6435 & 36 & 2036 & 31.64 \\
31 & satimage-2 & 5803 & 36 & 71 & 1.22 \\
32 & shuttle & 49097 & 9 & 3511 & 7.15 \\
33 & skin & 245057 & 3 & 50859 & 20.75 \\
34 & smtp & 95156 & 3 & 30 & 0.03 \\
37 & Stamps & 340 & 9 & 31 & 9.12 \\
38 & thyroid & 3772 & 6 & 93 & 2.47 \\
39 & vertebral & 240 & 6 & 30 & 12.50 \\
40 & vowels & 1456 & 12 & 50 & 3.43 \\
41 & Waveform & 3443 & 21 & 100 & 2.90 \\
42 & WBC & 223 & 9 & 10 & 4.48 \\
43 & WDBC & 367 & 30 & 10 & 2.72 \\
44 & Wilt & 4819 & 5 & 257 & 5.33 \\
45 & wine & 129 & 13 & 10 & 7.75 \\
46 & WPBC & 198 & 33 & 47 & 23.74 \\
47 & yeast & 1484 & 8 & 507 & 34.16 \\
\bottomrule
\end{tabular}
\end{table}

\FloatBarrier

\section{Statistics of synthetic Gaussian mixture datasets}

In Tables~\ref{tab:statistics_local}, \ref{tab:statistics_global}, and \ref{tab:statistics_cluster}, we report summary statistics for the synthetic Gaussian mixture datasets with respectively local, global, and cluster anomalies. The tables contain columns analogous to those in Table~\ref{app:statics_real}, except for the dataset name.

\begin{table}[!ht]
\centering
\scriptsize
\caption{Statistics of Gaussian mixture datasets with \emph{local} anomalies.}
\label{tab:statistics_local}
\begin{tabular}{lcccc}
\toprule
Id & \#Samples & \#Features & \#Anomalies & \%Anomalies \\
\midrule
0 & 5272 & 21 & 492 & 9.33 \\
1 & 6033 & 24 & 494 & 8.19 \\
2 & 5611 & 49 & 495 & 8.82 \\
3 & 8136 & 26 & 494 & 6.07 \\
4 & 8048 & 43 & 496 & 6.16 \\
5 & 3302 & 38 & 494 & 14.96 \\
6 & 4317 & 28 & 492 & 11.40 \\
7 & 9849 & 21 & 497 & 5.05 \\
8 & 7525 & 49 & 492 & 6.54 \\
9 & 8626 & 16 & 499 & 5.78 \\
10 & 7303 & 44 & 496 & 6.79 \\
11 & 3375 & 50 & 496 & 14.70 \\
12 & 6684 & 5 & 490 & 7.33 \\
13 & 5984 & 21 & 495 & 8.27 \\
14 & 5481 & 27 & 487 & 8.89 \\
15 & 4519 & 24 & 496 & 10.98 \\
16 & 7187 & 34 & 494 & 6.87 \\
17 & 8657 & 47 & 493 & 5.69 \\
18 & 9888 & 48 & 495 & 5.01 \\
19 & 8015 & 12 & 491 & 6.13 \\
20 & 5976 & 28 & 493 & 8.25 \\
21 & 6355 & 36 & 495 & 7.79 \\
22 & 8259 & 12 & 492 & 5.96 \\
23 & 6358 & 27 & 498 & 7.83 \\
24 & 8578 & 24 & 492 & 5.74 \\
\bottomrule
\end{tabular}
\end{table}

\begin{table}[!ht]
\centering
\scriptsize
\caption{Statistics of Gaussian mixture datasets used in experiments with \emph{global} anomalies.}
\label{tab:statistics_global}
\begin{tabular}{lcccc}
\toprule
Id & \#Samples & \#Features & \#Anomalies & \%Anomalies \\
\midrule
0 & 3885 & 11 & 496 & 12.77 \\
1 & 3921 & 49 & 499 & 12.73 \\
2 & 7330 & 27 & 490 & 6.68 \\
3 & 3290 & 30 & 496 & 15.08 \\
4 & 4271 & 14 & 496 & 11.61 \\
5 & 4228 & 40 & 493 & 11.66 \\
6 & 8118 & 27 & 490 & 6.04 \\
7 & 4789 & 28 & 496 & 10.36 \\
8 & 6979 & 4 & 491 & 7.04 \\
9 & 4791 & 12 & 496 & 10.35 \\
10 & 4288 & 31 & 497 & 11.59 \\
11 & 3130 & 50 & 498 & 15.91 \\
12 & 2810 & 21 & 497 & 17.69 \\
13 & 9375 & 9 & 491 & 5.24 \\
14 & 9579 & 7 & 499 & 5.21 \\
15 & 8560 & 15 & 496 & 5.79 \\
16 & 2695 & 48 & 493 & 18.29 \\
17 & 5527 & 41 & 493 & 8.92 \\
18 & 6272 & 20 & 493 & 7.86 \\
19 & 3274 & 17 & 497 & 15.18 \\
20 & 3016 & 30 & 489 & 16.21 \\
21 & 3218 & 35 & 498 & 15.48 \\
22 & 6339 & 4 & 491 & 7.75 \\
23 & 3131 & 40 & 495 & 15.81 \\
24 & 9868 & 22 & 494 & 5.01 \\
\bottomrule
\end{tabular}
\end{table}

\begin{table}[!ht]
\centering
\scriptsize
\caption{Overview of dataset statistics for \emph{cluster} anomalies.}
\label{tab:statistics_cluster}
\begin{tabular}{lcccc}
\toprule
Id & \#Samples & \#Features & \#Anomalies & \%Anomalies \\
\midrule
0 & 5908 & 7 & 496 & 8.40 \\
1 & 6257 & 15 & 498 & 7.96 \\
2 & 2823 & 22 & 498 & 17.64 \\
3 & 6691 & 31 & 494 & 7.38 \\
4 & 3680 & 32 & 491 & 13.34 \\
5 & 7791 & 7 & 497 & 6.38 \\
6 & 2675 & 30 & 496 & 18.54 \\
7 & 4277 & 29 & 495 & 11.57 \\
8 & 7240 & 27 & 491 & 6.78 \\
9 & 5092 & 3 & 495 & 9.72 \\
10 & 9584 & 4 & 498 & 5.20 \\
11 & 2847 & 35 & 497 & 17.46 \\
12 & 4793 & 20 & 490 & 10.22 \\
13 & 8916 & 25 & 496 & 5.56 \\
14 & 7344 & 35 & 499 & 6.79 \\
15 & 9607 & 29 & 496 & 5.16 \\
16 & 8115 & 27 & 490 & 6.04 \\
17 & 2661 & 6 & 491 & 18.45 \\
18 & 3884 & 3 & 497 & 12.80 \\
19 & 5124 & 2 & 498 & 9.72 \\
20 & 2755 & 40 & 491 & 17.82 \\
21 & 6501 & 43 & 499 & 7.68 \\
22 & 5618 & 15 & 491 & 8.74 \\
23 & 5493 & 27 & 498 & 9.07 \\
24 & 2889 & 24 & 495 & 17.13 \\
\bottomrule
\end{tabular}
\end{table}

\section{Extended details of the experimental setup}

Our experiments rely on the official code repository of the ADBench benchmark, which is publicly available at \href{https://github.com/Minqi824/ADBench}{https://github.com/Minqi824/ADBench}. For TCCM, we utilized \href{https://github.com/ZhongLIFR/TCCM-NIPS}{https://github.com/ZhongLIFR/TCCM-NIPS}, and for uTabPFN, we used the TabPFN extensions repository \linebreak \href{https://github.com/PriorLabs/tabpfn-extensions}{https://github.com/PriorLabs/tabpfn-extensions}.

For the experiment described in Section 4.1 of the main paper, each dataset was split into context (train) and query (test) sets in a 7:3 ratio when the proportion of anomalies was $\leq30\%$, and in a 6:4 ratio otherwise. For Sections 4.2 and 4.4, we employed the \emph{train\_test\_split} function from the \href{https://scikit-learn.org/stable/}{scikit-learn} library with the \emph{stratify} parameter set to \emph{labels}, dividing datasets in a 7:3 ratio. For Section 4.3, the results for the anomaly level equal to 0\% were copied from Section 4.1, while for higher anomaly levels, we first applied the stratification pipeline from Section 4.2 and then reduced the context set to match the exact anomaly levels of 5, 10, 15, and 20\%. Anomalous points removed from the context were completely excluded from evaluation and were not included in the query set. The ablation study in Section 4.5 follows the strategy from Section 4.1 for the clean-context scenario, and the strategy from Section 4.2 for the noisy-context setup.

\section{Extended experimental results}

In this section, we report the results on all datasets considered in this paper. 

\subsection{Can We Pretrain for Real Anomaly Detection?}
\label{app:clean}

Table~\ref{app:clean_aucroc} and Table~\ref{app:clean_f1} show the performance of all methods on our benchmark datasets, reporting AUCROC and F1 metrics averaged over 5 random seeds under the clean-context scenario. Each table  at the bottom contains a summary, presenting the mean metric score across datasets (Mean), together with the median method ranking (Median-Rank).

\begin{table}[!ht]
\centering
\tiny
\caption{Detailed AUCROC performance comparison across all datasets in the \emph{clean-context} scenario.}
\hspace*{-2.125cm}
\label{app:clean_aucroc}
\begin{tabular}{lccccccccccccccccc}
\toprule
Id & \our{} & uTabPFN & KNN & LOF & TCCM & IForest & PCA & HBOS & OCSVM & CBLOF & DeepSVDD & ECOD & COPOD & LODA & DAGMM & SOD & COF \\
\midrule
1 & 61.58 & 55.98 & 63.05 & \textbf{66.84} & 57.11 & 55.76 & 56.34 & 54.49 & 55.17 & 54.63 & 55.46 & 55.21 & 53.36 & 51.06 & 51.86 & 60.28 & \underline{63.28} \\
2 & 88.29 & \textbf{94.64} & 87.04 & 90.92 & 76.64 & \underline{91.44} & 81.23 & 70.78 & 66.03 & 63.83 & 85.28 & 78.97 & 77.56 & 60.61 & 60.87 & 71.39 & 64.30 \\
4 & 99.37 & 75.75 & 99.49 & 72.64 & \textbf{99.56} & \underline{99.54} & 99.16 & 99.49 & 99.43 & 99.22 & 98.20 & 98.97 & 99.50 & 98.48 & 95.54 & 84.55 & 45.03 \\
6 & 92.86 & 89.90 & 95.79 & 95.24 & \textbf{98.11} & 94.85 & 96.70 & 86.82 & \underline{97.45} & 87.88 & 93.77 & 93.41 & 92.28 & 92.22 & 86.06 & 50.32 & 52.91 \\
7 & 71.52 & 80.12 & 79.78 & 80.12 & 81.57 & 80.05 & \underline{82.24} & 70.35 & \textbf{84.27} & 79.11 & 79.47 & 78.07 & 66.82 & 78.70 & 74.37 & 36.06 & 41.14 \\
8 & 74.72 & 57.90 & 65.04 & 45.99 & 69.26 & 70.48 & \textbf{78.87} & \underline{75.94} & 70.16 & 69.21 & 72.55 & 75.11 & 74.55 & 64.03 & 52.58 & 49.33 & 40.94 \\
10 & 94.18 & \underline{98.61} & 97.60 & \textbf{99.06} & 91.91 & 84.65 & 94.44 & 76.68 & 95.21 & 89.93 & 87.64 & 91.56 & 87.90 & 92.47 & 87.51 & 59.56 & 51.55 \\
\bottomrule
\end{tabular}
\end{table}

\begin{table}[!ht]
\centering
\scriptsize
\caption{Per-dataset F1 assessment in the \emph{clean-context} setting.}
\hspace*{-1.75cm}
\label{app:clean_f1}
\begin{tabular}{lcccccccccccccc}
\toprule
Id & \our{} & KNN & LOF & IForest & PCA & HBOS & OCSVM & CBLOF & DeepSVDD & ECOD & COPOD & LODA & SOD & COF \\
\midrule
1 & 19.96 & \underline{20.93} & \textbf{23.85} & 11.56 & 14.11 & 13.06 & 13.78 & 14.67 & 13.62 & 9.99 & 9.10 & 11.81 & 20.14 & 19.19 \\
2 & 63.13 & 59.36 & \textbf{70.91} & \underline{67.01} & 56.61 & 51.13 & 38.22 & 38.79 & 60.39 & 44.75 & 37.69 & 33.03 & 49.19 & 23.74 \\
4 & 98.91 & \textbf{99.02} & 0.00 & 98.87 & \underline{98.93} & 98.84 & 98.90 & 98.57 & 98.58 & 93.68 & 94.74 & 98.84 & 87.21 & 0.00 \\
6 & 73.08 & \underline{85.15} & 84.26 & 79.94 & 84.74 & 65.78 & \textbf{87.05} & 66.41 & 80.11 & 72.46 & 68.37 & 78.03 & 30.67 & 28.14 \\
7 & 54.04 & \underline{65.87} & 63.09 & 59.93 & 65.20 & 43.64 & \textbf{68.65} & 58.53 & 62.24 & 46.16 & 36.88 & 63.19 & 32.37 & 20.70 \\
8 & 25.99 & 12.91 & 4.27 & 27.80 & \textbf{38.74} & \underline{33.60} & 29.11 & 29.24 & 27.50 & 33.45 & 32.24 & 14.75 & 11.03 & 2.89 \\
10 & \textbf{42.98} & 38.80 & \underline{39.40} & 17.56 & 35.19 & 16.21 & 36.18 & 33.62 & 22.17 & 29.91 & 23.08 & 30.97 & 9.08 & 6.91 \\
\bottomrule
\end{tabular}
\end{table}

\FloatBarrier

\subsection{How Robust Is ICL to Noisy Contexts?}

Similar as in Appendix~\ref{app:clean}, Tables~\ref{app:noisy_aucroc} and~\ref{app:noisy_f1} provide the full AUCROC and F1 performance averaged across 5 seeds, but for the \emph{noisy-context} scenario.

\begin{table}[!ht]
\centering
\tiny
\caption{Examination of AUCROC performance across all datasets under the \emph{noisy-context} setting.}
\hspace*{-2.125cm}
\label{app:noisy_aucroc}
\begin{tabular}{lccccccccccccccccc}
\toprule
Id & \our{} & uTabPFN & KNN & LOF & TCCM & IForest & PCA & HBOS & OCSVM & CBLOF & DeepSVDD & ECOD & COPOD & LODA & DAGMM & SOD & COF \\
\midrule
1 & 59.86 & 53.69 & 62.55 & \textbf{66.95} & 57.37 & 55.93 & 56.08 & 54.04 & 54.90 & 50.54 & 55.21 & 55.48 & 53.20 & 50.64 & 51.31 & 62.07 & \underline{64.11} \\
2 & 76.20 & \underline{88.49} & 71.72 & 70.19 & \textbf{92.68} & 81.99 & 64.94 & 59.88 & 58.04 & 50.63 & 68.27 & 78.53 & 76.95 & 46.86 & 55.46 & 77.92 & 67.30 \\
4 & 95.00 & 95.48 & 97.37 & 41.00 & \textbf{99.56} & 98.33 & 96.32 & 98.67 & 91.81 & 77.91 & 90.33 & 99.21 & \underline{99.55} & 98.53 & 93.42 & 93.55 & 38.67 \\
6 & 87.94 & 57.39 & 75.61 & 66.45 & 67.28 & 92.71 & \textbf{95.31} & 86.53 & 93.32 & 72.94 & 88.74 & \underline{93.47} & 92.43 & 88.25 & 79.24 & 70.31 & 64.67 \\
7 & 63.87 & 41.43 & 55.95 & 59.75 & 78.37 & 68.71 & 75.89 & 60.17 & \underline{78.52} & 61.08 & 73.60 & \textbf{78.81} & 66.99 & 73.48 & 65.16 & 51.57 & 53.25 \\
8 & \textbf{85.36} & 59.40 & 60.33 & 40.76 & 67.03 & 69.88 & \underline{78.26} & 75.03 & 68.67 & 56.34 & 71.00 & 75.52 & 74.75 & 61.92 & 51.66 & 48.54 & 42.00 \\
10 & \textbf{98.09} & 85.88 & 87.07 & 85.94 & \underline{94.56} & 87.10 & 93.75 & 77.15 & 91.84 & 62.54 & 86.32 & 91.51 & 88.17 & 88.08 & 83.01 & 75.09 & 79.80 \\
\bottomrule
\end{tabular}
\end{table}

\begin{table}[!ht]
\centering
\scriptsize
\caption{F1 score comparison for all datasets evaluated in the \emph{noisy-context} scenario.}
\hspace*{-1.75cm}
\label{app:noisy_f1}
\begin{tabular}{lcccccccccccccc}
\toprule
Id & \our{} & KNN & LOF & IForest & PCA & HBOS & OCSVM & CBLOF & DeepSVDD & ECOD & COPOD & LODA & SOD & COF \\
\midrule
1 & 7.94 & 10.73 & \textbf{12.55} & 5.45 & 7.10 & 5.93 & 6.84 & 4.98 & 6.69 & 4.69 & 4.49 & 6.19 & 9.58 & \underline{11.15} \\
2 & 27.24 & 24.37 & 24.15 & \textbf{31.70} & 22.17 & 23.95 & 14.65 & 8.31 & 24.01 & \underline{28.52} & 22.81 & 10.33 & 27.79 & 22.62 \\
4 & \textbf{84.78} & 37.96 & 0.00 & 52.13 & 45.34 & 46.31 & 46.26 & 47.13 & 44.05 & 47.90 & 47.17 & 47.23 & \underline{58.70} & 0.64 \\
6 & 40.58 & 38.42 & 24.87 & 50.33 & \textbf{64.41} & 51.22 & \underline{60.28} & 42.01 & 50.24 & 53.00 & 57.41 & 50.61 & 29.10 & 24.70 \\
7 & \underline{40.08} & 27.99 & 22.55 & 31.85 & 35.14 & 32.37 & \textbf{41.04} & 32.70 & 35.25 & 35.61 & 30.15 & 38.19 & 28.65 & 19.72 \\
8 & 10.80 & 4.68 & 1.57 & 12.95 & \textbf{18.58} & 16.75 & 13.67 & 7.86 & 12.66 & \underline{16.99} & 16.30 & 6.00 & 4.19 & 1.34 \\
10 & 5.63 & 11.15 & 10.41 & 8.39 & \textbf{13.83} & 6.47 & \underline{12.29} & 3.20 & 8.32 & 11.54 & 9.53 & 9.84 & 5.06 & 8.10 \\
\bottomrule
\end{tabular}
\end{table}

\FloatBarrier

\subsection{How Does Context Corruption Affect Performance?}

In Tables~\ref{tab:anom_level_auroc} and~\ref{tab:anom_level_f1}, we show the numerical values of AUCROC and F1 used for plotting Figure~6 in the main text.

\begin{table}[!ht]
\centering
\caption{Average AUCROC across different anomaly levels in the context set.}
\label{tab:anom_level_auroc}
\begin{tabular}{lccccccc}
\toprule
Anomaly level & IForest & KNN & PCA & TACTIC-N & TACTIC-C & uTabPFN \\
\midrule
0 & 82.34 & 88.44 & 78.85 & 87.17 & 90.43 & 88.90 \\
5 & 75.80 & 77.03 & 70.25 & 80.26 & 75.61 & 79.00 \\
10 & 69.07 & 72.66 & 63.50 & 74.81 & 70.83 & 69.47 \\
15 & 69.88 & 71.50 & 63.89 & 74.45 & 68.65 & 67.46 \\
20 & 68.01 & 71.25 & 61.63 & 71.54 & 64.64 & 67.41 \\
\bottomrule
\end{tabular}
\end{table}

\begin{table}[!ht]
\centering
\caption{Mean F1 performance under varying anomaly levels in the context set.}
\label{tab:anom_level_f1}
\begin{tabular}{lcccccc}
\toprule
Anomaly level & IForest & KNN & PCA & TACTIC-N & TACTIC-C \\
\midrule
0 & 47.74 & 58.24 & 44.30 & 55.61 & 65.52 \\
5 & 38.75 & 39.33 & 36.22 & 47.65 & 36.16 \\
10 & 36.65 & 39.71 & 33.48 & 51.25 & 38.76 \\
15 & 35.52 & 35.66 & 33.54 & 50.12 & 34.66 \\
20 & 32.94 & 33.41 & 30.43 & 48.10 & 33.09 \\
\bottomrule
\end{tabular}
\end{table}

\FloatBarrier

\subsection{Can One Model Handle Multiple Anomaly Types?}

Tables~\ref{tab:local_aucroc}, \ref{tab:local_f1}, \ref{tab:global_aucroc}, \ref{tab:global_f1}, \ref{tab:cluster_aucroc}, \ref{tab:cluster_f1} contain detailed AUCROC and F1 performance results for experiments on Gaussian mixture datasets featuring respectively local, global, and cluster anomalies.

\begin{table}[!ht]
\centering
\tiny
\caption{AUCROC evaluation on synthetic Gaussian mixture datasets with \emph{local} anomalies.}
\label{tab:local_aucroc}
\begin{tabular}{lcccccccccccc}
\toprule
Id & \our{} & uTabPFN & KNN & LOF & TCCM & IForest & PCA & HBOS & OCSVM & CBLOF & COPOD & COF \\
\midrule
0 & \textbf{99.52} & 99.35 & \textbf{99.52} & 99.50 & 83.80 & 99.21 & 99.39 & 99.06 & 99.38 & 75.12 & 98.71 & 98.48 \\
1 & 99.63 & 99.66 & \underline{99.68} & \textbf{99.72} & 86.28 & 99.39 & 99.59 & 99.36 & 99.54 & 74.88 & 99.01 & 98.51 \\
2 & \underline{99.99} & \underline{99.99} & \underline{99.99} & \textbf{100.00} & 95.79 & 99.96 & \underline{99.99} & 99.97 & \underline{99.99} & 64.35 & 99.89 & 99.68 \\
3 & \underline{99.92} & 99.87 & \underline{99.92} & \textbf{99.95} & 90.22 & 99.68 & 99.87 & 99.75 & 99.84 & 75.72 & 99.49 & 99.21 \\
4 & 99.96 & \underline{99.98} & \underline{99.98} & \textbf{99.99} & 92.66 & 99.92 & 99.96 & 99.94 & 99.96 & 66.99 & 99.86 & 99.66 \\
5 & \textbf{99.99} & 99.97 & \textbf{99.99} & \textbf{99.99} & 90.44 & 99.95 & 99.98 & 99.93 & 99.98 & 71.85 & 99.87 & 99.58 \\
6 & 99.85 & 99.83 & \textbf{99.88} & \textbf{99.88} & 88.19 & 99.62 & 99.84 & 99.70 & 99.83 & 75.02 & 99.46 & 99.31 \\
\bottomrule
\end{tabular}
\end{table}

\begin{table}[!ht]
\centering
\scriptsize
\caption{Method comparison using F1 on synthetic Gaussian mixture datasets with \emph{local} anomalies.}
\label{tab:local_f1}
\begin{tabular}{lcccccccccc}
\toprule
Id & \our{} & KNN & LOF & IForest & PCA & HBOS & OCSVM & CBLOF & COPOD & COF \\
\midrule
0 & 86.79 & \textbf{90.92} & \underline{90.46} & 88.66 & 89.93 & 88.61 & 90.05 & 37.65 & 86.24 & 84.90 \\
1 & \textbf{87.06} & \underline{84.64} & 84.60 & 82.72 & 84.42 & 83.58 & 83.68 & 33.98 & 81.97 & 79.68 \\
2 & \textbf{98.01} & 92.03 & \underline{92.96} & 92.93 & 92.30 & 92.52 & 92.70 & 29.04 & 92.29 & 90.73 \\
3 & \textbf{89.36} & 73.91 & \underline{74.18} & 72.88 & 73.59 & 73.65 & 73.36 & 30.10 & 72.98 & 73.12 \\
4 & \textbf{95.31} & 75.47 & 75.98 & 74.88 & 75.65 & 74.64 & 76.36 & 25.40 & 73.74 & \underline{79.14} \\
5 & \textbf{98.27} & 78.64 & 78.26 & \underline{79.87} & 78.17 & 79.54 & 77.67 & 44.59 & 77.73 & 74.45 \\
6 & \textbf{94.11} & \underline{92.84} & 92.60 & 92.23 & 92.76 & 91.86 & 91.84 & 41.42 & 90.83 & 86.73 \\
\bottomrule
\end{tabular}
\end{table}

\begin{table}[!ht]
\centering
\tiny
\caption{AUCROC performance benchmarking on synthetic Gaussian mixture datasets with \emph{global} anomalies.}
\label{tab:global_aucroc}
\begin{tabular}{lcccccccccccc}
\toprule
Id & \our{} & uTabPFN & KNN & LOF & TCCM & IForest & PCA & HBOS & OCSVM & CBLOF & COPOD & COF \\
\midrule
0 & \underline{99.95} & 99.68 & \textbf{99.96} & 99.94 & 85.98 & 99.87 & 99.88 & 99.83 & 99.90 & 73.68 & 99.56 & 99.70 \\
1 & \textbf{100.00} & \textbf{100.00} & \textbf{100.00} & \textbf{100.00} & 99.50 & \textbf{100.00} & \textbf{100.00} & \textbf{100.00} & \textbf{100.00} & 59.97 & \textbf{100.00} & 99.99 \\
2 & \textbf{100.00} & 99.99 & \textbf{100.00} & \textbf{100.00} & 99.38 & \textbf{100.00} & \textbf{100.00} & \textbf{100.00} & \textbf{100.00} & 64.92 & \textbf{100.00} & 99.99 \\
3 & \textbf{100.00} & \textbf{100.00} & \textbf{100.00} & \textbf{100.00} & 96.57 & \textbf{100.00} & \textbf{100.00} & \textbf{100.00} & \textbf{100.00} & 67.66 & \textbf{100.00} & 99.93 \\
4 & \textbf{99.97} & 99.75 & \textbf{99.97} & \textbf{99.97} & 90.42 & 99.85 & 99.92 & 99.81 & 99.92 & 72.68 & 99.69 & 99.69 \\
5 & \textbf{100.00} & \textbf{100.00} & \textbf{100.00} & \textbf{100.00} & 99.37 & \textbf{100.00} & \textbf{100.00} & \textbf{100.00} & \textbf{100.00} & 59.17 & \textbf{100.00} & 99.97 \\
6 & \textbf{100.00} & 99.99 & \textbf{100.00} & \textbf{100.00} & 99.53 & \textbf{100.00} & \textbf{100.00} & \textbf{100.00} & \textbf{100.00} & 64.61 & \textbf{100.00} & 99.97 \\
\bottomrule
\end{tabular}
\end{table}

\begin{table}[!ht]
\centering
\scriptsize
\caption{F1-based evaluation on synthetic Gaussian mixture datasets with \emph{global} anomalies.}
\label{tab:global_f1}
\begin{tabular}{lcccccccccc}
\toprule
Id & \our{} & KNN & LOF & IForest & PCA & HBOS & OCSVM & CBLOF & COPOD & COF \\
\midrule
0 & 87.09 & 89.33 & 89.17 & 89.33 & \underline{89.49} & 88.79 & \textbf{89.58} & 45.59 & 87.99 & 85.82 \\
1 & \textbf{100.00} & 86.89 & 87.58 & 88.02 & 86.47 & 87.87 & 86.48 & 42.40 & \underline{88.88} & 79.32 \\
2 & \textbf{94.84} & 78.51 & 79.40 & 79.24 & 79.34 & 79.55 & 79.08 & 27.23 & \underline{79.91} & 79.49 \\
3 & \textbf{99.40} & 78.70 & 79.22 & 76.43 & 78.01 & 79.22 & 78.09 & 53.27 & \underline{79.82} & 66.22 \\
4 & 89.99 & 93.60 & \underline{93.74} & \textbf{93.86} & 93.68 & 93.68 & 93.46 & 43.70 & 91.85 & 88.81 \\
5 & \textbf{99.40} & 90.28 & 91.24 & 91.74 & 90.75 & \underline{92.30} & 91.00 & 37.78 & 91.66 & 85.51 \\
6 & \textbf{93.12} & 74.33 & 73.53 & 74.57 & 74.51 & 74.37 & 74.61 & 26.67 & 74.23 & \underline{76.02} \\
\bottomrule
\end{tabular}
\end{table}

\begin{table}[!ht]
\centering
\tiny
\caption{AUCROC performance analysis on Gaussian mixture synthetic datasets with \emph{cluster} anomalies.}
\label{tab:cluster_aucroc}
\begin{tabular}{lcccccccccccc}
\toprule
Id & \our{} & uTabPFN & KNN & LOF & TCCM & IForest & PCA & HBOS & OCSVM & CBLOF & COPOD & COF \\
\midrule
0 & \textbf{89.00} & 79.86 & \underline{82.74} & 78.87 & 51.85 & 82.43 & 81.77 & 80.56 & 82.33 & 55.26 & 81.24 & 76.60 \\
1 & \textbf{96.30} & \underline{74.71} & 72.41 & 70.06 & 53.53 & 70.44 & 69.62 & 69.20 & 70.43 & 58.16 & 71.18 & 66.90 \\
2 & \textbf{81.04} & 58.71 & 66.62 & 63.90 & 55.03 & 68.12 & 69.20 & 68.01 & 68.41 & 56.68 & \underline{69.82} & 64.84 \\
3 & \textbf{70.73} & 67.27 & 69.43 & 68.54 & 59.96 & 69.30 & \underline{70.05} & 69.81 & 69.83 & 53.20 & 68.40 & 65.56 \\
4 & \underline{69.76} & \textbf{69.82} & 68.71 & 67.97 & 61.45 & 68.61 & 69.66 & 69.13 & 69.47 & 59.45 & 68.32 & 67.13 \\
5 & \textbf{87.98} & 80.91 & \underline{85.09} & 81.20 & 63.30 & 83.85 & 84.21 & 83.05 & 83.59 & 59.31 & 80.20 & 79.99 \\
6 & \textbf{70.15} & 59.79 & 62.08 & 61.22 & 55.92 & 65.67 & 66.70 & 66.59 & 66.27 & 56.61 & \underline{67.99} & 59.60 \\
\bottomrule
\end{tabular}
\end{table}

\begin{table}[!ht]
\centering
\scriptsize
\caption{Comparison of F1 scores on synthetic Gaussian mixture datasets with \emph{cluster} anomalies.}
\label{tab:cluster_f1}
\begin{tabular}{lcccccccccc}
\toprule
Id & \our{} & KNN & LOF & IForest & PCA & HBOS & OCSVM & CBLOF & COPOD & COF \\
\midrule
0 & \textbf{37.76} & 32.71 & 26.51 & 32.48 & 32.34 & 32.02 & 34.06 & 10.56 & \underline{36.29} & 28.99 \\
1 & \textbf{44.50} & 23.93 & 20.24 & 23.46 & 24.31 & 24.22 & 23.71 & 17.44 & \underline{27.49} & 17.83 \\
2 & \textbf{48.97} & 21.84 & 18.65 & 23.71 & 24.01 & 24.26 & 23.84 & 20.66 & \underline{25.96} & 19.52 \\
3 & \textbf{21.42} & 17.39 & 16.36 & \underline{19.68} & 18.62 & 17.97 & 18.31 & 14.49 & 18.37 & 13.97 \\
4 & \textbf{31.79} & 26.23 & 25.54 & 25.68 & 27.77 & 28.16 & \underline{28.38} & 23.27 & 26.83 & 20.83 \\
5 & 28.67 & \underline{35.42} & 28.36 & \textbf{35.53} & 33.97 & 33.65 & 32.88 & 12.95 & 29.42 & 28.29 \\
6 & \textbf{39.92} & 19.05 & 19.68 & 22.33 & 21.64 & 21.54 & 21.73 & 21.33 & \underline{24.66} & 15.12 \\
\bottomrule
\end{tabular}
\end{table}

\FloatBarrier

\subsection{Which Priors Matter for Pretraining?}

Tables~\ref{tab:ablation_aucroc} and~\ref{tab:ablation_f1} present the impact of the priors used for pretraining on AUROC and F1. We consider both clean- and noisy-context scenarios.

\begin{table}[!ht]
\centering
\scriptsize
\caption{AUCROC of TACTIC variants pretrained on various priors in clean- and noisy-context setups.}
\label{tab:ablation_aucroc}
\begin{tabular}{lccc|ccc}
\toprule
& & Clean & & & Noisy & \\
Id & \our{} & w/o GMM & w/o PFN & \our{} & w/o GMM & w/o PFN \\
\midrule
1 & 61.58 & 56.89 & 50.00 & 59.86 & 57.01 & 55.41 \\
2 & 88.29 & 83.52 & 91.94 & 76.20 & 77.55 & 67.40 \\
4 & 99.37 & 99.46 & 61.44 & 95.00 & 92.80 & 55.97 \\
6 & 92.86 & 96.91 & 89.74 & 87.94 & 85.34 & 64.05 \\
7 & 71.52 & 81.02 & 68.41 & 63.87 & 60.23 & 52.63 \\
8 & 74.72 & 59.91 & 50.00 & 85.36 & 81.40 & 39.90 \\
\bottomrule
\end{tabular}
\end{table}

\begin{table}[!ht]
\centering
\scriptsize
\caption{F1 of TACTIC variants pretrained on various priors in clean- and noisy-context setups.}
\label{tab:ablation_f1}
\begin{tabular}{lccc|ccc}
\toprule
& & Clean & & & Noisy & \\
Id & \our{} & w/o GMM & w/o PFN & \our{} & w/o GMM & w/o PFN \\
\midrule
1 & 19.96 & 16.63 & 18.86 & 7.94 & 8.17 & 3.74 \\
2 & 63.13 & 52.33 & 62.57 & 27.24 & 24.80 & 24.83 \\
4 & 98.91 & 98.87 & 14.55 & 84.78 & 86.72 & 4.75 \\
6 & 73.08 & 84.93 & 1.57 & 40.58 & 34.98 & 9.61 \\
7 & 54.04 & 54.80 & 0.00 & 40.08 & 33.97 & 17.06 \\
8 & 25.99 & 14.51 & 13.87 & 10.80 & 10.13 & 0.40 \\
\bottomrule
\end{tabular}
\end{table}

\end{document}